%% file: sample-base.tex
\documentclass{article}
\usepackage[utf8]{inputenc}
\usepackage{graphicx}
\usepackage{hyperref}
\usepackage{amsmath}
\usepackage{amssymb}
\usepackage{caption}
\usepackage{subcaption}
\usepackage{xcolor}
\usepackage{hyperref}
\usepackage{bm}
\usepackage{algorithm2e}
\usepackage{geometry}
\usepackage{forest}
\usepackage{hyperref}

\usetikzlibrary{shadows}
\usepackage{csvsimple}
\DeclareMathOperator*{\argmin}{arg\,min}

\usepackage{multirow}
\usepackage{soul}
\usepackage{amsfonts}
\usepackage{tikz}
\usetikzlibrary{trees}
\usepackage{booktabs}

\newcommand{\botrule}{\bottomrule}

\newcommand{\dd}{\mathop{}\mathopen{}\mathrm{d}}

\usepackage{arydshln}

\usepackage{tikz}
\tikzset{
    ncbar angle/.initial=90,
    ncbar/.style={
        to path=(\tikztostart)
        -- ($(\tikztostart)!#1!\pgfkeysvalueof{/tikz/ncbar angle}:(\tikztotarget)$)
        -- ($(\tikztotarget)!($(\tikztostart)!#1!\pgfkeysvalueof{/tikz/ncbar angle}:(\tikztotarget)$)!\pgfkeysvalueof{/tikz/ncbar angle}:(\tikztostart)$)
        -- (\tikztotarget)
    },
    ncbar/.default=0.2cm,
}
\tikzset{square left brace/.style={ncbar=0.1cm}}
\tikzset{square right brace/.style={ncbar=-0.1cm}}

\title{Interpretable Scientific Discovery with Symbolic Regression: A Review}
\author{Nour Makke\; Sanjay Chawla}
\date{
    Qatar Computing Research Institute, HBKU, Doha\\[2ex]
    \today
}

\begin{document}

\maketitle

\begin{abstract}
Symbolic regression is emerging as a promising machine learning method for learning succinct underlying interpretable mathematical expressions directly from data. Whereas it has been traditionally tackled with genetic programming, it has recently gained a growing interest in deep learning as a data-driven model discovery method, achieving significant advances in various application domains ranging from fundamental to applied sciences. In this survey, we present a structured and comprehensive overview of symbolic regression methods and discuss their strengths and limitations. 
\end{abstract}

\maketitle
\setlength\parindent{0pt}

\newpage
\tableofcontents
\newpage


\input intro.tex
\input probdef.tex
\input srmethods.tex
\input linearsr.tex
\input nonlinearsr.tex

\input tree.tex
\input srapplications.tex
\input datasets.tex
\input conclusion.tex
\bibliographystyle{ieeetr}

\bibliography{sample-base}

\newpage
\input appendix.tex

\end{document}

%% file: intro.tex
\section{Introduction}

Symbolic Regression (SR) is a rapidly growing subfield within machine learning (ML) to
infer symbolic mathematical expressions from data~\cite{koza,doi:10.1126/science.1165893}. Interest in SR is being driven by the observation that it 
is not sufficient to only have accurate predictive models; however, it is often necessary that the learned
models be interpretable~\cite{rudin}. A model
is interpretable if the relationship 
between the input and output of the model
can be logically or mathematically traced in a succinct manner.
In other words, learnable models are interpretable if expressed as mathematical equations.
As ``disciplines'' become increasingly data-rich and adopt
ML techniques, the demand for interpretable
models is likely to grow. For example, in the natural sciences (e.g., physics), mathematical models derived from first principles make it possible to reason about the underlying phenomenon in a way that is not possible with predictive models like deep neural networks. In critical disciplines like healthcare, non-interpretable models may never be allowed to be deployed - however accurate they maybe~\cite{healthcare}. \\

\noindent
{\bf Example:} Consider a data set consisting of samples $(q_1,q_2,r,F)$, where $q_1$ and $q_2$ are the charges of two particles, $r$ is the distance between them and $F$ is the measured force between the particles. Assume $q_1, q_2$, and $r$ are the input variables, and $F$ is the output
variable. Suppose we model the input-output relationship as $F = \theta_0 + \theta_1q_1 + \theta_2q_2 + \theta_3r$. 
Then, using the data set, we can infer the model's parameters ($\theta_i$). The model
will be interpretable because we will know the impact of each variable on the output. For example,
if $\theta_3$ is negative, then that implies that as $r$ increases, the force $F$ will decrease. From physics,
we know that this form of the model is unlikely to be accurate. On the other hand, we could model the relationship
using a neural network $F = NN(q_1,q_2,r,\theta)$. We expect the model to be highly accurate and predictive because neural networks (NNs) are universal function approximators. However, the model is uninterpretable because the input-output relationship is not easily apparent.
The input feature vector subsequently undergoes several layers of nonlinear transformations, i.e., $y = \sigma(\sum_iW_i~\sigma(\sum_jW_j~\sigma(\sum_kW_k~\sigma(\cdots \sum_{\ell}W_{\ell}\rm{x}))))$, where $\sigma$ is a nonlinear activation function, and $W_{idx}$ are the learnable parameters of NN layer of index $idx$. Such models, called {\em ``blackbox"}, do not have an internal logic to let users understand how inputs are mathematically mapped to outputs. Explainability is the application of other methods to explain model predictions and to understand how it is learned. It refers to why the model makes the decision that way. What distinguishes explainability from interpretability is that interpretable models are transparent~\cite{rudin}. For example, the linear regression model predictions can be interpreted by evaluating the relative contribution of individual features to the predictions using their weights.
An ideal SR model will return the relationship
as $k\frac{q_1q_2}{r^2}$, which is the definition of the Coulomb force between two charged particles with a constant\footnote{$k$ is the electric force (or Coulomb) constant, $k= 8.9875517923\times10^9$ kg$\cdot$m$^{3}\cdot$s$^{-4}\cdot$A$^{-2}$ in SI base units.} $k = 8.98 \times 10^{9}$. However, learning the SR model is highly non-trivial as it involves searching over a large space of
mathematical operations and identifying the right constant ($k$) that will fit the data. SR models can be directly inferred from data or can be used to {\em ``whitebox"} a {\em ``blackbox"} model such as a neural network.\\

The ultimate goal of SR is to bridge data and observations following the Keplerian trial and error approach~\cite{kepler}. 
Kepler developed a data-driven model for planetary motion 
using the most accurate astronomical measurements of the era, which resulted in elliptic orbits described by a power law. In contrast, Newton developed a dynamic relationship between physical variables that described the underlying process at the origin of these elliptic orbits. Newton's approach~\cite{newton} led to three laws of motion later verified by experimental observations. Whereas both methods fit the data well, Newton's approach could be generalized to predict behavior in regimes where no data were available. Although SR is regarded as a data-driven model discovery tool, it aims to find a symbolic model that simultaneously fits data well and could be generalized to uncovered regimes.\\

SR is deployed as an interpretable and predictive ML model or a data-driven scientific discovery method.
SR was investigated as early as 1970 in research works~\cite{gerwin,langley,falkenhainer} aiming to rediscover empirical laws. Such works iteratively apply a set of data-driven heuristics to formulate mathematical expressions.
The first AI system meant to automate scientific discovery is called BACON~\cite{bacon1, 10.5555/29379}. It was developed by Patrick Langley in the late 1970s and was successful in rediscovering versions of various physical laws, such as 
Coulomb's law and Galileo's laws for the pendulum and constant acceleration, among many others.
SR was later studied by Koza~\cite{10.5555/1623755.1623877,10.5555/892491,koza} who proposed that genetic programming (GP) can be used to discover symbolic models by encoding mathematical expressions as computational trees, where GP is an evolutionary algorithm that iteratively evolves an initial population of individuals via biology-inspired operations. SR was since then tackled with GP-based methods~\cite{koza,keijer,vladislavleva,Korns2011,Uy2010SemanticallybasedCI,jin,dsr,ffx,gpasr,ITEA,mrgp,lacavagp,livingreview}.
Moreover, it was popularized as a data-driven scientific discovery tool with the commercial software Eureqa~\cite{Dubcakova:2011:GPEM} based on a research work~\cite{doi:10.1126/science.1165893}. Whereas GP-based methods achieve high prediction accuracy, they do not scale to high dimensional data sets and are sensitive to hyperparameters~\cite{dsr}.
More recently, SR has been addressed with deep learning-based methods \cite{Udrescu:2019mnk,DBLP:journals/corr/MartiusL16,dsr,dsrgp,metamodel,transformers,srthatscales,sindyae}
%
which leverage neural networks (NNs) to learn accurate symbolic models.
%
%
SR has been applied in fundamental and applied sciences such as  astrophysics~\cite{cranmersolar}, chemistry~\cite{chemistry,hernandez2019fast}, materials science~\cite{Wang_2019,Weng2020SimpleDD}, semantic similarity measurement~\cite{MARTINEZGIL2020113663}, climatology~\cite{Abdellaoui2021SymbolicRF}, medicine~\cite{10.1117/1.JMI.7.4.046501}, among many  others. Many of these applications are promising, showing the potential of SR.
A recent SR benchmarking platform \textit{SRBench} is introduced by La Cava \textit{et al.}~\cite{srbench}. It comprises 14 SR methods (among which ten are GP-based), applied on 252 data sets. The goal of \textit{SRBench} was to provide a benchmark for rigorous evaluation and comparison of SR methods.\\

This survey aims to help researchers effectively and comprehensively understand the SR problem and how it could be solved, as well as to present the current status of the advances made in this growing subfield. 
The survey is structured as follows. First, we define the SR problem, present a structured and comprehensive review of methods, and discuss their strengths and limitations. Furthermore, we discuss the adoption of these SR methods across various application domains and assess their effectiveness. 
Along with this survey, a living review~\cite{livingreview} aims to group state-of-the-art SR methods and applications and track advances made in the SR field. The objective is to update this list often to incorporate new research works.\\

This paper is organized as follows. The SR problem definition is presented in Section~\ref{probdef}. We present an overview of methods deployed to solve the SR problem in Section~\ref{sec:srmethods}, and the methods are discussed in detail in Sections~\ref{sec:linear},~\ref{sec:nonlinear} and ~\ref{sec:tree_expression}. Selected applications are described and discussed in Section~\ref{sec:srapplications}.
Section~\ref{sec:data} presents an overview of existing benchmark data sets. Finally, we summarize our conclusions and discuss perspectives in Section~\ref{sec:conclusion}.

%% file: probdef.tex
\section{Problem Definition}\label{probdef}

The problem of symbolic regression can be defined in terms of classical Empirical Risk Minimization (ERM)~\cite{ERM}.\\

{\bf Data:} Given a data set $\mathcal{D} = \{(\mathbf{x}_i,y_i)\}_{i=1}^{n}$, where $\mathbf{x}_i \in \mathbb{R}^{d}$ is the input vector
and $y_{i} \in \mathbb{R}$ is a scalar output. \\

{\bf Function Class:} Let $\mathcal{F}$ be a function class consisting of mappings $f: \mathbb{R}^{d} \rightarrow \mathbb{R}$. \\

{\bf Loss Function:} Define the loss function for every candidate $f \in \mathcal{F}$: 

\begin{equation}
    l(f):= \sum_{i=1}^{n} l(f(x_{i}),y_{i})
\end{equation}

A common choice is the squared difference between the output and prediction, i.e. $l(f) = \sum_i (y_i-f(x_i))^2$.\\

{\bf Optimization:} The optimization task is to find the function ($f$) over the set of functions $\mathcal{F}$ that minimizes the loss function:

\begin{equation}
   f^{*} = \argmin_{f \in\mathcal{F}} l(f)
\end{equation}

As stated below, what distinguishes SR from conventional regression problems is the discrete nature of the function class $\mathcal{F}$. Different methods for solving the SR problem reduce to characterizing the function class. 

\subsection{Class of Function}\label{subsec:functionclass}

In SR, to define $\mathcal{F}$, we specify 
a library of elementary arithmetic operations and mathematical functions and variables, and an element
$f \in \mathcal{F}$ is the set of all functions that can be obtained by function composition in the library~\cite{srnphard}. For example, consider
a library:
\begin{equation}\label{eq:libraryL}
L = \{\mathrm{id}(\cdot),~\mathrm{add}(\cdot,\cdot),~\mathrm{sub}(\cdot,\cdot),~\mathrm{mul}(\cdot,\cdot),+1,-1\}
\end{equation}

Then the set of all of the polynomials (in one variable $x$)  with integer coefficients can be derived
from $L$ using function composition.


\subsection{Expression representation}
\label{subsec:functionrepresentation}

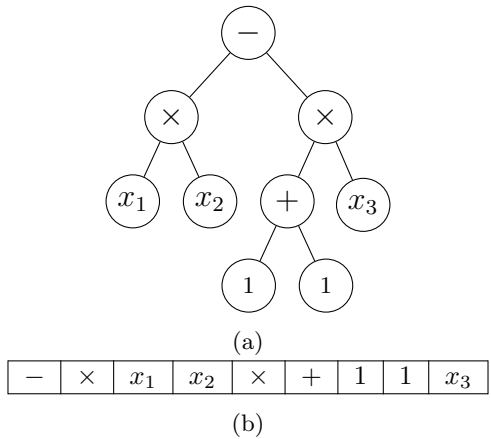
\begin{figure}[h]
    \centering
    \begin{subfigure}[b]{0.5\textwidth}
        \centering
        \begin{forest}
        for tree = {circle, draw,
                minimum size=2.em,
                inner sep=0.5pt,
                font=\large,
                l sep=4mm,s sep=3mm
                }
        [$-$[$\times$[$x_1$][$x_2$]][$\times$[$+$[\small 1][\small 1]][$x_3$]]]
        \end{forest}%
        \caption{}
        \label{fig:tree_subfig1}
        \end{subfigure}
    \hfill
    \begin{subfigure}[b]{0.5\textwidth}
    \centering
    \begin{tabular}{|c|c|c|c|c|c|c|c|c|c|}\hline
    $-$ & $\times$ & $x_1$ & $x_2$ & $\times$ & $+$ & $1$ & $1$ & $x_3$   \\ \hline
    \end{tabular}
    \caption{}
    \label{fig:tree_suubfig2}
    \end{subfigure}
    \caption{(a) Example of a unary-binary tree  that encodes $f(\mathrm{x}) = x_1x_2 - 2x_3$. (b) Sequence representation of the tree-like structure of $f(\mathrm{x})$.}
\end{figure}

It is convenient to express symbolic
expressions in a sequential form using either a unary-binary expression
tree or the polish notation~\cite{polishnotation}. For example, the expression $f(\mathrm{x}) = x_1x_2 - 2x_3$ can be derived using function
composition from $L$ (Eq.~\ref{eq:libraryL}) and represented as a tree-like structure illustrated in Figure~\ref{fig:tree_subfig1}.
By traversing the (binary) tree top to bottom and left to right in a depth-first manner, we can represent the same expression as a unique sequence called the polish form, as illustrated in Figure~\ref{fig:tree_suubfig2}.


In practice, the library $L$ includes many other common elementary mathematical functions, including the basic trigonometric functions like sine, cosine, logarithm, exponential, square root, power low, etc.
Prior domain knowledge is advantageous for library definition because it reduces the search space to only include the most relevant mathematical operations to the studied problem.
Furthermore, a large
range of possible numeric constants should
be possible to express. For example, numbers in base-10 floating point notation rounded up to four
significant digits can be represented as triple of ({\em sign}, {\em mantissa}, {\em exponent})~\cite{transformers}. The function $\sin(3.456x)$, for example,
can be represented as $[\sin,~\mathrm{mul},~3456,~E-3,~x]$.

%% file: srmethods.tex
\section{Symbolic regression methods overview}
\label{sec:srmethods}

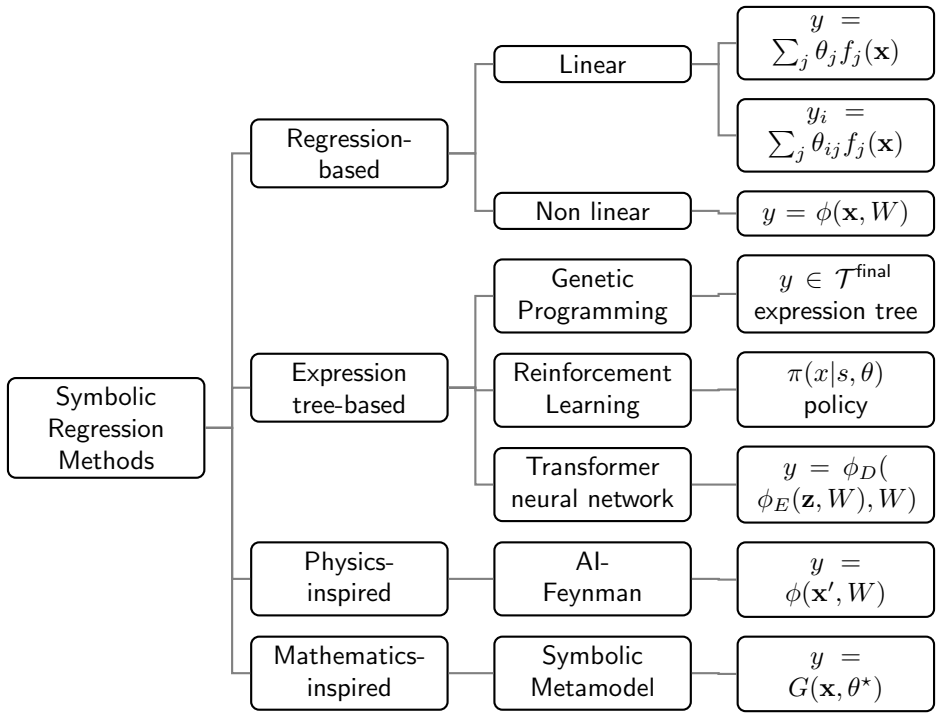
\begin{figure}[htp]
  \centering  
\begin{forest}
    for tree={%
        my node,
        where level=4{text width=2em}{},
        l sep+=5pt,
        grow'=east,
        edge={gray, thick},
        parent anchor=east,
        child anchor=west,
        if n children=0{tier=last}{},
        edge path={
            \noexpand\path [draw, \forestoption{edge}] (!u.parent anchor) -- +(10pt,0) |- (.child anchor)\forestoption{edge label};
        },
        if={isodd(n_children())}{
            for children={
                if={equal(n,(n_children("!u")+1)/2)}{calign with current}{}
            }
        }{}
    }
    [Symbolic Regression Methods
    [Regression-based
    [Linear[{$y=\sum_j\theta_jf_j(\mathbf{x})$}
    ][{$y_i=\sum_j\theta_{ij}f_j(\mathbf{x})$}
    ]]
    [Non linear[{\vphantom{N}{$y=\phi(\mathbf{x},W)$}}
    ]]]
    [Expression tree-based
    [Genetic Programming[{\vphantom{G}{$y\in\mathcal{T}^{\text{final}}$} \\ expression tree\vphantom{g}}
    ]]
    [Reinforcement Learning[{\vphantom{T}{$\pi(x|s,\theta)$}\\ policy}
    ]]
    [Transformer neural network\vphantom{q}
    [{\vphantom{T}{$y = \phi_{D}($} \\{$\left.\phi_{E}(\mathbf{z},\small W),W\right)$}}
    ]]]
    [Physics-\\inspired
    [AI-\\Feynman[{\vphantom{A}{$y = $} \\ {$\phi(\mathbf{x}^{\prime},W)$}}
    ]]]
    [Mathematics-inspired
    [Symbolic Metamodel\vphantom{p}[{\vphantom{P}{$y=$}\\{$ G(\mathbf{x},\theta^{\star})$}}
    ]]]
    ]
  \end{forest}\\[5pt]
  \caption{Taxonomy based on the type of symbolic regression methods. $\phi$ denotes a neural network function, $W$ denotes the set of learnable parameters in NN. $\mathbf{x}$ denotes the input data, $\mathbf{z}$ denotes a reduced representation of $\mathbf{x}$, and $\mathbf{x}^{\prime}$ denotes a new representation of $\mathbf{x}$, e.g., by defining new features based on the original ones. $\mathcal{T}$ represents the final population of selected expression trees in genetic programming.}
  \label{fig:srtaxonomy}
\end{figure}

In this survey, we categorize SR methods in the following manner: regression-based methods, expression tree-based methods, physics-inspired and mathematics-inspired methods, as presented in  Figure~\ref{fig:srtaxonomy}. For each category, a summary of the mathematical tool, the expression form, the set of unknowns, and the search space, is presented in Table~\ref{tab:srmethods}.
\\

\newbox\mybox

\setbox\mybox=\hbox{%
  \begin{forest}
    for tree={circle,draw}
    [a[b][c]]
  \end{forest}%
}
\tikzset{
    my node/.style={
        align=center,
        draw=black,
        inner color=white,
        outer color=white,
        thick,
        minimum width=0.9cm,
        rounded corners=3,
        text width=2.35cm,
        font=\sffamily,
    }
}

The linear method defines the functional form as a linear combination of nonlinear functions of $x$ that are comprised in the predefined library $L$. Linear models are expressed as:

\begin{equation}
    f(\mathrm{x},\theta) = \sum_{j}\theta_j h_j(\mathrm{x})
\end{equation}

where $j$ spans the base functions of $L$. 
The optimization problem reduces to find the set of parameters $\{\theta\}$ that minimizes the loss function defined over a continuous parameter space $\Theta = \mathbb{R}^{M}$ as follows:

\begin{equation}
    \theta^{*} = \argmin_{\theta\in\Theta} ~\sum_il(f(x_i,\theta),y_i)
\end{equation}

This method is advantageous for being deterministic and disadvantageous because it imposes a single model structure which is fixed during training when the model's parameters are learned.\\

The nonlinear method defines the model structure by a neural network. Nonlinear models can thus be expressed as:

\begin{equation}
    f(\mathrm{x},W) = \sigma(\sum_iW_i~\sigma(\sum_jW_j~\sigma(\cdots \sum_{\ell}W_{\ell}\rm{x}))))
\end{equation}

where $\sigma$ is a nonlinear activation function, and $W_{idx}$ are the learnable parameters of NN layer of index $idx$.
Similarly to the linear method, the optimization problem reduces finding the set of parameters $\{W,b\}$ of neural network layers, which minimizes the loss function over the space of real values.\\

Expression tree-based methods treat mathematical expressions as unary-binary trees whose internal nodes are operators and terminals are operands (variables or constants). This category comprises GP-based, deep neural transformers, and reinforcement learning-based methods. 
In GP-based methods, a set of transition rules (e.g., mutation, crossover) is defined over the tree space and applied to an initial population of trees throughout many iterations until the loss function is minimized. Transformers~\cite{DBLP:journals/corr/VaswaniSPUJGKP17} represent a novel architecture of neural network (encoder and decoder) that uses attention mechanism. The latter was primarily used to capture long-range dependencies in a sentence. Transformers were designed to operate on sequential data and to perform sequence-to-sequence (seq2seq) tasks. For their use in SR, input data points $(\mathbf{x},y)$ and symbolic expressions ($f$) are encoded as sequences and transformers perform set-to-sequence tasks. The unknowns are the weight parameters of the encoder and the decoder.
Reinforcement learning (RL) is a machine learning method that seeks to learn a policy $\pi(x|\theta)$ by training an agent to perform a task by interacting with its environment in discrete time steps. An RL setting requires four components: state space, action space, state transition probabilities, and reward. The agent selects an action that is sent to the environment. A reward and a new state are sent back to the agent from its environment and used by the agent to improve its policy at the next time step. In the context of SR, symbolic expression (sequence) represents a state, predicting an element in a sequence represents an action, the parent and sibling represent the environment, and the reward is commonly chosen as the mean square error (MSE). RL-based SR methods are commonly hybrid and use various ML tools (e.g., NN, RNN, etc.) in a joint manner with RL. 

\begin{table}[h]
\begin{center}
\caption{Table summarizing symbolic regression methods. The mathematical tool, the expression form, the set of unknowns, and the search space are specified for each method. Set2seq abbreviates ``set-to-sequence".}
\label{tab:srmethods}
\begin{tabular}{@{}lllll@{}}
\toprule
Method & Tool & Expression form & Unkown & Search space \\
\midrule
\multirow{2}{*}{Linear SR} & Uni-D linear system
&$y = \sum_i\theta_if_i(\mathbf{x})$ & 
$\{\theta\}_{i}$ & $\mathbb{R}$ \\[5pt]
& Multi-D linear system & ${y}_i = \sum_j\theta_jf_j(\mathbf{x})$ & $(\{\theta\}_{i})_{j}$ & $\mathbb{R}$\\[5pt] \cmidrule{2-5}
Nonlinear SR & Neural Network & $y=f(W\cdot\mathbf{x}+b)$ & $\{W,b\}$ & $\mathbb{R}$\\[5pt]
\midrule
\multirow{3}{*}{Expression-tree search} 
    & Genetic Programming & Expression tree & & trees\\[5pt]
    & Transformers & set2seq mapping & $\{W_q,W_k,W_v\}$ & $\mathbb{R}$\\[5pt]
    & Reinforcement learning & set2seq mapping & $\pi(\theta)$ & $\mathbb{R}$ \\[5pt]
\midrule
Physics-inspired & AI-Feynman & $y= f(\mathbf{x},\theta)$ &  $-$ & $-$ \\[5pt]
\midrule
Mathematics-inspired & Symbolic metamodels & $G(\mathbf{x},\theta)$ & $\theta$ & $\mathbb{R}$\\ 
\botrule
\end{tabular}
\end{center}
\end{table}

%% file: linearsr.tex
\section{Linear symbolic regression}
\label{sec:linear}

The linear approach assumes, by definition, that the target symbolic expression ($f(x)$) is a linear combination of nonlinear functions of feature attributes: 

\begin{equation}\label{linearSRfunction}
 f(\mathrm{x}) = \sum_j \theta_j h_j(\mathrm{x})
\end{equation}

\noindent
Here $\mathrm{x}$ denotes the input features vector, $\theta_j$ denotes a weight coefficient, and $h_j(\cdot)$ denotes a unary operator of the library $L$. This approach predefines the model's structure and reduces the SR problem to learn only the model's parameters by solving a system of linear equations. The particular case where $f(x)$ is a linear combination of degree-one monomial reduces to a conventional linear regression problem, i.e., $f(x) = \sum_j \theta_jx^j = \theta_0 + \theta_1 x + \theta_2 x^2 +\cdots$. There exist two cases for this problem: (1) a unidimensional case defined by $f: \mathbb{R}^{d}\rightarrow\mathbb{R}$; and (2) a multidimensional case defined by $f: \mathbb{R}^{d}\rightarrow\mathbb{R}^{m}$, with $d$ the number of input features and $m$ the number of variables required for a complete description of a system; for example, the Lorenz system for fluid flow is defined in terms of three physical variables which depend on time. 

\subsection{Unidimensional case}
\label{subsec:unid}

\noindent
Given a data set $\mathcal{D}= \{(x_i,y_i)\}_{i=1}^{n}$, the mathematical expression could be either univariate ($x_i\in\mathbb{R},~ y_i=f(x_i)$) or multivariate ($\mathbf{x}_i\in\mathbb{R}^{d},~ y_i=f(\mathbf{x}_i)$). The methodology of linear SR is presented in detail for the univariate case in Secion~\ref{subsec:univariate} for simplicity and is extended for the multivariate case in Section~\ref{subsec:multivariate}.

\subsubsection{Univariate function}
\label{subsec:univariate}

{\bf Data set:} $\mathcal{D}=\{x_i\in\mathbb{R};~y_i=f(x_i)\}$. \\

{\bf Library:} $L$ can include any number of mathematical operators such that the dimension of the data set is always greater than the dimension of the library matrix (see discussion below).\\

In this approach, a coefficient $\theta_j$ is assigned to each candidate function ($f_j(\cdot)\in L$) as an activeness criterion such that:

\begin{equation}\label{eq:linear_uni}
    y = \sum_j \theta_j f_j(x)
\end{equation}

Applying Eq.~\ref{eq:linear_uni} to input-output pairs $(x_i,y_i)$ yields a system of linear equations as follows:

\begin{equation}
    \begin{matrix}
        y_1 = \theta_0 +~\theta_1f_1(x_1) +~\theta_2f_2(x_1) +~\cdots +~\theta_kf_k(x_1)\\
        y_2 = \theta_0 +~\theta_1f_1(x_2) +~\theta_2f_2(x_2) +~\cdots +~\theta_kf_k(x_2)\\
        \vdots\\
        y_n = \theta_0 +~\theta_1f_1(x_n) +~\theta_2f_2(x_n) +~\cdots +~\theta_kf_k(x_n)\\
    \end{matrix}
\end{equation}
\\

which can be represented in a matrix form as:

\begin{equation}\label{eq:linearsystem}
    \left[\begin{matrix}
        y_1 \\ y_2\\ \vdots\\ y_n
    \end{matrix}\right] = 
    \left[\begin{matrix}
        1 & f_1(x_1) & f_2(x_1) & \cdots & f_k(x_1)\\
        1 & f_1(x_2) & f_2(x_2) & \cdots & f_k(x_2)\\
        \vdots \\
        1 & f_1(x_n) & f_2(x_n) & \cdots & f_k(x_n)
    \end{matrix}\right]
    \left[\begin{matrix}
        \theta_0 \\ \theta_1\\ \vdots\\ \theta_k
    \end{matrix}\right]
\end{equation}
\\

Equation~\ref{eq:linearsystem} can then be presented in a compact form:

\begin{equation}
    \label{eq:yutheta}
    \mathrm{Y} = \mathrm{U}(\mathrm{X})\cdot\rm{\mathrm{\Theta}}
\end{equation}

where $\mathrm{\Theta} \in \mathbb{R}^{(k+1)}$ is the sparse vector of coefficients, and $\mathrm{U} \in \mathbb{R}^{n\times(k+1)}$ is the library matrix which can be represented as a function of the input vector $\mathrm{X}$ as follows:

\begin{equation}\label{umatrix}
\mathrm{U}(\mathrm{X}) = 
\left[~
\begin{matrix}
\mid \quad & \mid \quad & \mid \quad & & \mid\\
\mathrm{1} \quad & f_1(\mathrm{X}) \quad & f_2(\mathrm{X}) \quad  & \cdots & f_k(\mathrm{X})\\
\mid \quad & \mid \quad & \mid \quad & &  \mid\\
\end{matrix}
\right]
\end{equation}

{\bf Example:} For a library defined as:

\begin{equation}\label{eq:library_example}
    L = \{1,~x,~(\cdot)^2,~\sin(\cdot),~\cos(\cdot),~\exp(\cdot)\}
\end{equation}
\\

The matrix $\mathrm{U}$ becomes:

\[
\mathrm{U}(\mathrm{X}) = 
\left[~
\begin{matrix}
\mid \quad & \mid \quad & \mid \quad & \mid & \mid & \mid\\
\mathrm{1} \quad & \mathrm{X} \quad & \mathrm{X}^2 & \sin(\mathrm{X}) & \cos(\mathrm{X}) & \exp(\mathrm{X}) \\
\mid \quad & \mid \quad & \mid \quad & \mid & \mid & \mid\\
\end{matrix}
\right]
\]
\\\

Each row (of index $i$) in Eq.~\ref{umatrix} is a vector of $(k+1)$ functions of $x_{i}$. The vector of coefficients, i.e., the model's parameters, is obtained by solving Eq.~\ref{eq:yutheta} as follows:\footnote{Technically the pseudo-inverse, $U^{+}$}:

\begin{equation}
    \mathrm{\Theta} = (\mathrm{U}^{\rm{T}}\mathrm{U})^{-1}\mathrm{U}^{\rm{T}}\mathrm{Y}
\label{eq:SRresult}
\end{equation}

The magnitude of a coefficient $\theta_k$ effectively measures the size of the contribution of the associated function $f_k(\cdot)$ to the final prediction. Finally, the prediction vector $\hat{\mathrm{Y}}$ can be evaluated using Eq.~\ref{eq:yutheta}.
\\

An exemplary schematic is illustrated in Figure~\ref{fig:schematic1d} for the univariate function $f(x) = 1+\alpha x^3$. Only coefficients associated with functions $\{1, x^3\}$ of the library are non-zero, with values equal to 1 and $\alpha$, respectively.
\\

\begin{figure}[htp]
    \centering
    \begin{tikzpicture}[circed/.style ={circle,fill}]]
        \draw [black, thick] (0.4,0) to [square left brace ] (0.4,2);
        \draw [line width=2mm, red ] (0.5,0.05) -- (0.5,1.95); \node at (0.5, 2.25) {$\mathrm{Y}$};
        \draw [black, thick] (0.6,0) to [square right brace] (0.6,2);
        \node at (1.1, 1) {$=$};
        \draw [black, thick] (1.6,0) to [square left brace ] (1.6,2);
        \draw [line width=2mm, gray ] (1.7,0.05) -- (1.7,1.95); \node at (1.7, 2.2) {$1$};
        \draw [line width=2mm, gray ] (2.1,0.05) -- (2.1,1.95); \node at (2.1, 2.2) {$x$};
        \draw [line width=2mm, gray ] (2.5,0.05) -- (2.5,1.95); \node at (2.55, 2.25) {$x^2$};
        \draw [line width=2mm, gray ] (2.9,0.05) -- (2.9,1.95); \node at (2.95, 2.25) {$x^3$};
        \node at (3.5, 1) {$\cdots$};
        \draw [black, thick] (3.7,0) to [square right brace] (3.7,2);
        \draw [black, thick] (4.1,0) to [square left brace ] (4.1,2);
        \draw [line width=2mm, gray ] (4.2,0.05) -- (4.2,1.95); \node at (4.2, 2.25) {$\mathrm{\Theta}$};
        \draw [black, thick] (4.3,0) to [square right brace] (4.3,2);
        \node [circed,fill=red, inner sep =1.5 pt] at (4.2, 1.85){};
        \node [circed,fill=red, inner sep =1.5 pt] at (4.2, 1.1){};
    \end{tikzpicture}
    \caption{Schematic of the system of linear equations of Eq.~\ref{eq:yutheta} for $f(x) = 1 + \alpha x^3$. A library matrix $\mathrm{U}(\mathrm{X})$ of nonlinear functions of the input is constructed, where $L = \{1,x,x^2,x^3, \cdots\}$. The marked entries in the $\mathrm{\Theta}$ vector denote the non-zero coefficients determining which functions of the library are active.}
    \label{fig:schematic1d}
\end{figure}


In the following, linear SR is tested on synthetic data. In each experiment, training and test data sets are generated. Each set consists of twenty data points randomly sampled from a uniform distribution $\mathrm{U}(-1,1)$, and $y$ is evaluated using a univariate function, i.e., $\mathcal{D}=\{(x_i,f(x_i))\}_{i=1}^{n}$. Two libraries are considered in these experiments: $L_1 = \{ x,(\cdot)^2,(\cdot)^3,\cdots,(\cdot)^9\}$ and $L_2 = L_1 \cup \{\sin(\cdot), \cos(\cdot), \tan(\cdot),  \exp(\cdot), \rm{sigmoid}(\cdot)\}$. The results are reported in terms of the output expression (Equation~\ref{linearSRfunction}) and the coefficient of determination $R^2$. SR problems are grouped into (i) pure polynomial functions and (ii) mixed polynomial and trigonometric functions. In each experiment, parameters are learned using the training data set, and results are reported for the test data set in Table~\ref{tab:results1d}.

\begin{table}[h]
\caption{Results of linear SR in the case of univariate functions. $\mathcal{D} = \{(x_i;y_i)\}$; $x_i\in\mathrm{U}(-1,1,20)$ and $y_i=f(x_i)$. $L_1 = \{x, (\cdot)^2,\cdots,(\cdot)^9\}$ and $L_2 = L_1 \cup \{\sin(\cdot),\cos(\cdot),\tan(\cdot),  \exp(\cdot),\mathrm{sigmoid}(\cdot)\}$.
T denotes True, and F denotes False.}
\label{tab:results1d}
\begin{center}
\begin{tabular}{@{}llllll@{}}
\toprule
Benchmark & Expression & \multicolumn{2}{c}{$L_1$} & \multicolumn{2}{c}{$L_2$} \\ \cmidrule{3-6}
& & Exp & $R^2$ & Exp  & $R^2$ \\
\midrule
    Nguyen-2 & $x^4 + x^3+ x^2 + x$& T & 1.0 & F & 0.886 \\
    Nguyen-3 & $x^5 + x^4 + x^3+ x^2 + x$& T & 1.0 & F & 0.867\\
    Livermore-21 & $x^8 + x^7 + x^6 + x^5 + x^4 + x^3+ x^2 + x$ &T & 1.0 & F & 0.869 \\
    Livermore-9 & $x^9 + x^8 + x^7 + x^6 + x^5 + x^4 + x^3+ x^2 + x$ &T & 1.0 & F & 0.882 \\
    Livermore-6 & $x + 2x^2 + 3x^3 + 4x^4$ & T & 1.0 & F & 0.417 \\
    Livermore-19 & $x + x^2 + x^4 + x^5$ & T & 1.0 & F & -0.079\\[3pt] \hline
    Livermore-14$^{*}$ & $x + x^2 + x^3 + \sin(x)$ & F & 1.0 & F & -0.857 \\
    Nguyen-5 & $\sin(x^2)\cos(x) -1$ & F & 0.999 & F & -3.97 \\
    Nguyen-6 & $\sin(x) + \sin(x+x^2)$ & F & 0.999 & F & 0.564\\ 
\botrule
\end{tabular}
\end{center}
\end{table}

\begin{figure}
\centering
    \includegraphics[width=0.55\textwidth]{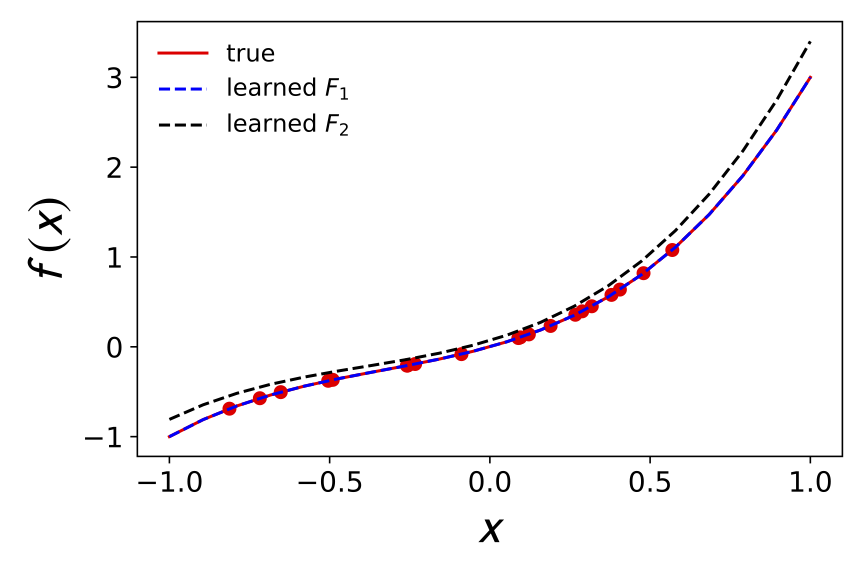}
    \label{fig:Ng1_linearSR_1d}
\caption{
Result of linear SR for the Nguyen-1 benchmark, i.e., $f(x) = x+x^2+x^3$. Red points represent (test) data set. The red curve represents the true function.  The blue and black dashed curves represent the learned functions using $L_1$ and $L_2$, respectively.} 
\label{fig:nguyen1}
\end{figure}

\noindent
For polynomial functions, an exact output is obtained using $L_1$ with an $R^2 = 1.0$, whereas only approximate output is obtained using $L_2$. In the latter case, the quality of the fit depends on the size of the training data set.
An exemplary result is shown in Figure~\ref{fig:nguyen1} for $f(x) = x + x^2 +x^3$. 
Points represent the (test) data of the input file, i.e., $\mathrm{X}$; the red curve represents $f(x)$ as a function of $x$, and the blue and black dashed curves represent the predicted function $\hat{f}(x)$ obtained using $L_1$ and $L_2$ respectively.
An exact match between the ground-truth function and the predicted one is found using $L_1$, whereas a significant discrepancy is obtained using $L_2$. This discrepancy could be explained by the fact that various functions in $L_2$ exhibit the same $x$-dependence over the covered  $x$-range.\\

For mixed polynomial and trigonometric expressions, 
both library choices do not produce the exact expression. However, a better $R^2$-coefficient is obtained using $L_1$. In the case of Nguyen-5 benchmark for example, i.e.,  $f(x) = \sin(x^2)\cos(x) -1$, the resulting function is the Taylor expansion of $f$:
$$\hat{y}(x) \approx -1 + 0.9x^2 -0.5x^4 - 0.13x^6 + \mathcal{O}(x^8)$$

In conclusion, this approach can not learn the ground-truth function when the latter is a multiplication of two functions (i.e., $f(x)=f_1(x)*f_2(x)$) or when it has a multiplicative or an additive factor to the variable (e.g., $\sin(\alpha + x),~\exp(\lambda*x)$, etc.). In the best case, it outputs an approximation of the ground-truth function. Furthermore, this approach fails to predict the correct mathematical expression when the library is extended to include a mixture of polynomial, trigonometric, exponential, and logarithmic functions.

\subsubsection{Multivariate function}
\label{subsec:multivariate}

For a given data set $\mathcal{D}=\{x_i\in\mathbb{R}^{d};~y_i=f(x_1,\cdots,x_d)\}$,
%
%
where $d$ is the number of features, 
the same equations presented in Section~\ref{subsec:univariate} are applicable. However, the dimension of the library matrix $\mathrm{U}$ changes to consider the features vector dimension. For example,  for the same library shown in Eq.~\ref{eq:library_example} and a two dimensional features vector, i.e., $\mathrm{X}\in\mathbb{R}^2$, $\mathrm{U}(\mathrm{X})$ becomes:

\begin{equation}\label{eq:umatrixmultivariate}
\begin{split}
\mathrm{U}(\mathrm{X}) &= 
\left[~
\begin{matrix}
\mid \quad & \mid \quad & \mid \quad & \mid & \mid & \mid\\
\mathrm{1} \quad & \mathrm{X} \quad & \mathrm{X}^{P2} & \sin(\mathrm{X}) & \cos(\mathrm{X}) & \exp(\mathrm{X}) \\
\mid \quad & \mid \quad & \mid \quad & \mid & \mid & \mid\\
\end{matrix}
\right]\\
 &= \left[~
 \begin{matrix}
 \mid & \mid & \mid & \mid & \mid & \mid & \mid & \mid & \\
 1 &\quad x_{1} & x_{2} &\quad x_{1}^2 & x_{1}x_2 & x_2^2 &\quad \sin(x_1) & \sin(x_2) &\quad \cdots\\
 \mid & \mid & \mid & \mid & \mid & \mid & \mid & \mid & \\
 \end{matrix}\right]
\end{split}
\end{equation}

Here, $\mathbf{X}^{P_q}$ denotes polynomials in $\mathrm{X}$ of the order $q$.
\\

Table~\ref{tab:results_2variables} presents the results of the experiments performed on two-variables dependent functions, i.e., $f(x_1,x_2)$. Similarly to Section~\ref{subsec:univariate}, training and test data sets are generated by randomly sampling twenty pairs of points ($x_1,x_2$) from a uniform distribution U(-1,1) such that $\mathcal{D}=\{(x_{1i},x_{2i},f(x_{1i},x_{2i}))\}_{i=1}^{n}$. The same choices for the library are considered: $L_1 = \{ x,(\cdot)^2,\cdots,(\cdot)^9\}$ and $L_2 = L_1 \cup \{\sin(\cdot), \cos(\cdot), \tan(\cdot),  \exp(\cdot), \rm{sigmoid}(\cdot)\}$. 
An exact match between the ground-truth and predicted function is obtained using $L_1$ for any polynomial function, whereas only approximate solutions are obtained for trigonometric functions. The results are approximate of the ground-truth function using $L_2$.

\begin{table}[h]
\caption{Results for multivariate functions using linear SR. $\mathcal{D} = \{(x_1,x_2)\in\mathrm{U}(-1,1,20);~y=f(x_1,x_2)\}$. $L_1 = \{ x,(\cdot)^2,\cdots,(\cdot)^9\}$ and $L_2 = L_1 \cup \{\sin(\cdot),\cos(\cdot),\tan(\cdot),  \exp(\cdot),\rm{sigmoid}(\cdot)\}$.
T and F refer to True and False.}
\label{tab:results_2variables}
\begin{center}
\begin{tabular}{@{}llllll@{}}
\toprule
Benchmark & Expression & \multicolumn{2}{c}{$L_1$} & \multicolumn{2}{c}{$L_2$} \\ \cmidrule{3-6}
& & Result & $R^2$ & Result  & $R^2$ \\
\midrule
    Nguyen-12 & $x_1^4 - x_1^3 + \frac{1}{2}x_2^2 - x_2$& T & 1.0 & F & $\approx 1$\\
    Livermore-5 & $x_1^4 - x_1^3 + x_1^2 -x_2$ & T & 1.0 & F & $\approx 1$\\
    Nguyen-9 & $\sin(x_1) + \sin(x_2^2)$ & F & $\approx 1$ & F & $\approx 1$\\
    Nguyen-10 & $2\sin(x_1)\cos(x_2)$ &  F & $\approx 1$ & F & $\approx 1$\\
\botrule
\end{tabular}
\end{center}
\end{table}

\begin{figure}[htp]
    \centering
    \includegraphics[width=.5\linewidth]{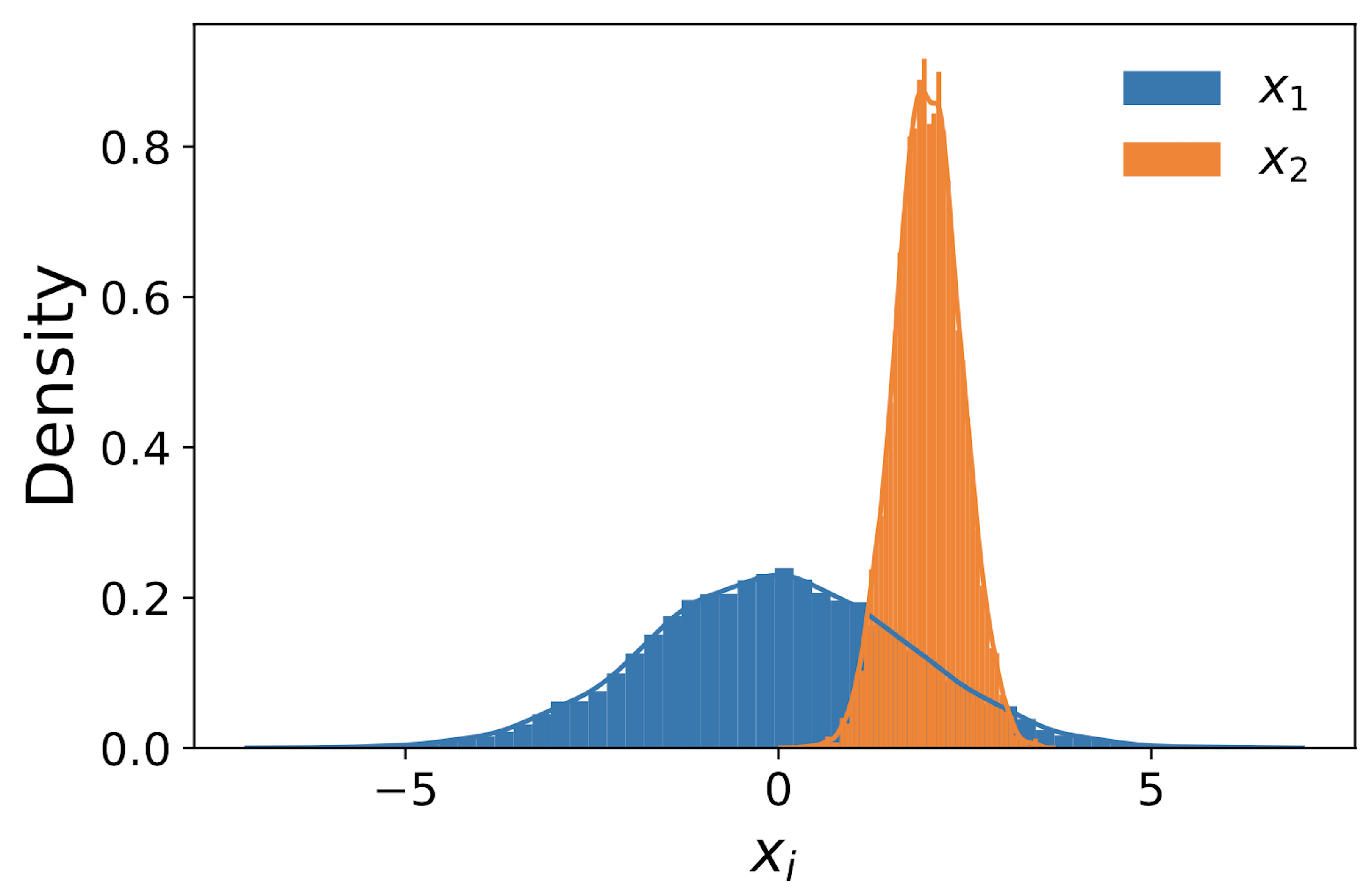}
    \caption{Two-dimensional multivariate normal distribution used in test applications.}
    \label{fig:dataset}
\end{figure}

Furthermore, linear SR is tested on a dataset generated using a two-dimensional multivariate normal distribution $\mathcal{N}(\mathbf{\mu},\mathbf{\Sigma})$, as shown in Fig.~\ref{fig:dataset}. Different analytic expressions for $f(x_1,x_2)$ were tested with different library bases that are summarized in Table~\ref{tab:2}, including pure polynomial basis functions, polynomial and trigonometric basis functions, and a mixed library.

\begin{table}[htp]
\begin{center}
\caption{Library bases used in test problems of Sec.~\ref{subsec:multivariate}}
\begin{tabular}{@{}lll@{}}
\toprule
& Name & List of functions \\ 
\midrule 
\multirow{3}{*}{Library}
& U1 & $\mathrm{1}, \mathrm{X}, \mathrm{X}^{P2}, \mathrm{X}^{P3},\mathrm{X}^{P4}$ \\ \cmidrule{2-3}
& U2 & $\mathrm{1}, \mathrm{X}, \mathrm{X}^{P2}, \cos(\mathrm{X}), \sin(\mathrm{X})$ \\ \cmidrule{2-3}
& U3 & $\mathrm{1}, \mathrm{X}, \mathrm{X}^{P2}, \cos(\mathrm{X}), \sin(\mathrm{X}), \tan(\mathrm{X}), \exp(\mathrm{X}), \mathrm{sigmoid}(\mathrm{X}$) \\
\botrule
\end{tabular}
\label{tab:2}
\end{center}
\end{table}

The function $y_1=\cos(x_1) + \sin(x_2)$ is explored with all three bases. In the case of a pure polynomial basis, the correct terms of the Taylor expansion of both $\cos(x_1)$ and $\sin(x_2)$ are identified with only approximate values of their coefficients, i.e., $\hat{y}_1 = (0.88 - 0.3x_1^2+0.01x_1^4) + (0.97-0.2x_2^3)$, which is reflected in the significantly high reconstruction error of the order of $30\%$. In both bases where trigonometric functions are enlisted, the correct terms $\cos(x_1)$ and $\sin(x_2)$ are identified with an excellent reconstruction error, that is $\ge 10^{-7}$. Note that the lowest reconstruction error is obtained for the library $\mbox{U}2$, which has the least number of operations and, consequently, the lowest number of coefficients.\\
%


The function $y_2 = x_1^2 + \cos(x_2)$ is also tested. For the pure polynomial basis, the reconstructed function $\hat{y}_2 = x_1^2 + (0.83+0.49x_2 -x_2^2)$ predicts approximate values with a reconstruction error of $\leq 1\%$. An excellent prediction is made for the other bases, which enlist both operations in $y_2(x_1,x_2)$.\\

\begin{figure}[htp]
    \centering
    \includegraphics[width=.9\linewidth]{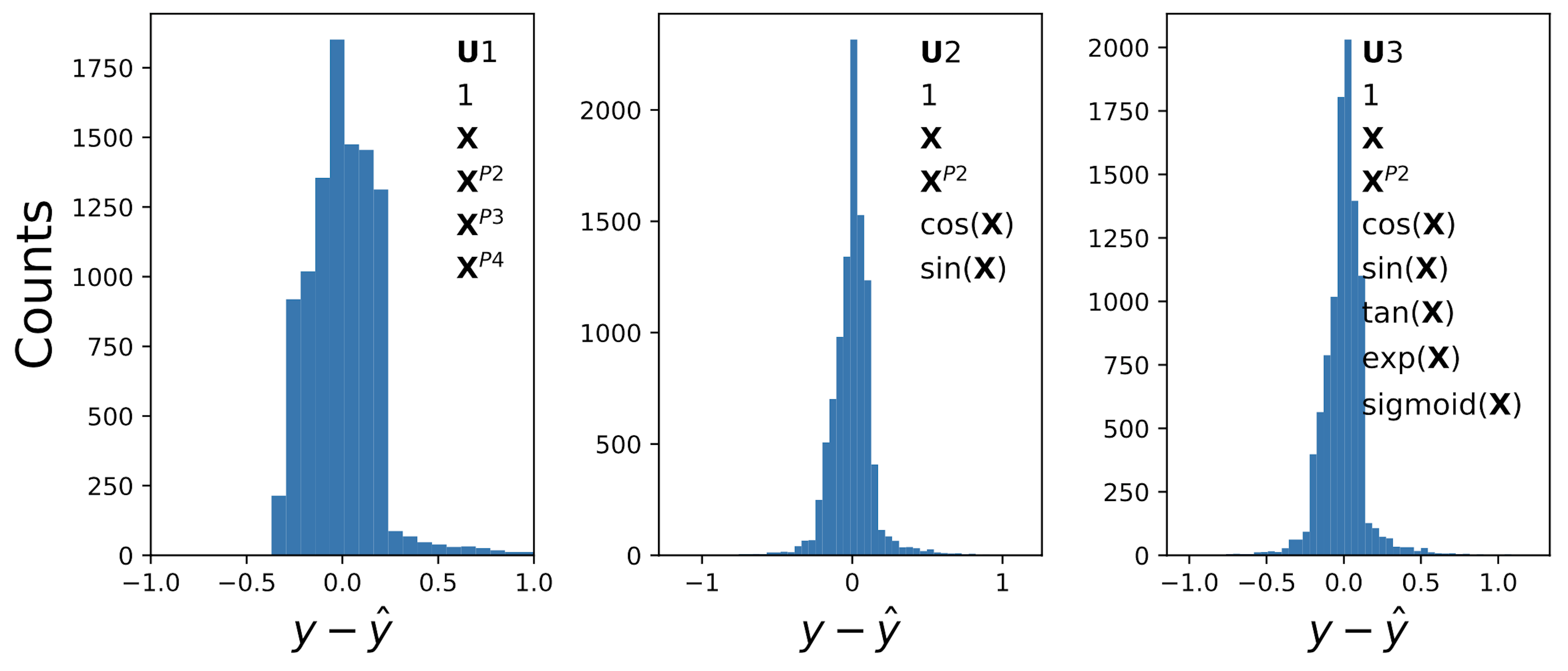}
    \caption{Difference between true ($y$) and predicted ($\hat{y}$) values of the function $y= x_1(1+x_2)+\cos(x_1)*\sin(x_2)$, for the three libraries defined in Table~\ref{tab:2}: $\mbox{U1}$~(left), $\mbox{U}2$~(center), $\mbox{U}3$~(right).}
    \label{fig:y3}
\end{figure}

In the same exercise, a more complicated function form is tested that includes mixed terms, i.e., $y_3 = x_1(1+x_2)+\cos(x_1)*\sin(x_2)$. The difference between the true and the predicted function is illustrated in Fig.~\ref{fig:y3}. The linear approach performs similarly for all three library bases. A low reconstruction error is obtained because the operation term $\cos(x_1)*\sin(x_2)$ in $y_3$ is not enlisted in any of the libraries, showing an important limitation of the current approach. 

\subsection{Multidimensional case}
\label{subsec:multid}

The target mathematical expression comprises $m$ components, i.e., $\mathrm{Y} = \left[ {y}_1,\cdots,{y}_m \right]$, and the goal is to learn the coefficients of a system of linear equations rather than one mathematical expression.
Each component ($y_j$) is described by:

\begin{equation}
\label{eq:roweq}
    \mathrm{y}_j = {f}_j(\mathrm{x}) = \sum_{k}\theta_{jk}h_{k}(\mathrm{x})
\end{equation}

In this case, there exist $m$ sparse vectors of coefficients, i.e., $\rm{\mathrm{\Theta}} = \left[ {\theta}_1~\cdots~{\theta}_m\right]$. Consider the Lorenz system, which is a set of ordinary differential equations that captures nonlinearities in the dynamics of fluid convection. It consists of three variables $\{x_1, x_2, x_3\}$ and their first-order derivatives with respect to time $\{\frac{\dd x_1}{\dd t},\frac{\dd x_2}{\dd t},\frac{\dd x_3}{\dd t}\}$, which we will refer to as $\{{y_1},{y_2},{y_3}\}$. Using the library of Eq.~\ref{eq:library_example}, the system of linear equations is represented in a matrix form as follows:

\begin{equation}\label{eq:linearsystem_multid}
    \left[\begin{matrix}
        y_{1} & y_{2} & y_{3}\\ 
        \vdots & \vdots & \vdots \\
        \vdots & \vdots & \vdots \\
    \end{matrix}\right] = 
    \left[\begin{matrix}
        1 & x_1 & x_2 & x_1^2 & x_1x_2 & x_2^2 & & \exp(x_2) \\
        \vdots & \vdots & \vdots & \vdots & \vdots & \vdots & \cdots & \vdots \\
        \vdots & \vdots & \vdots & \vdots & \vdots & \vdots &  & \vdots \\
    \end{matrix}\right]
    \left[\begin{matrix}
        \theta_{1} & \theta_{2} & \theta_{3}\\ 
        \vdots & \vdots & \vdots \\
        \vdots & \vdots & \vdots \\    \end{matrix}\right]
\end{equation}
\\

Here, $\mathrm{Y}\in\mathbb{R}^{n\times 3}$, $\mathrm{U}(\mathrm{X})\in\mathbb{R}^{n\times k}$ and ${\Theta}\in\mathbb{R}^{k\times 3}$, where $n$ is the size of the input data and $k$ is the number of columns in the library matrix $\mathrm{U}$. The $j^{th}$-component of the $\mathrm{Y}$ vector is given by:

\begin{equation}
    y_j = \theta_{j,0} + \theta_{j,1} x_1 + \theta_{j,2} x_2 + \theta_{j,3} x_1^2 + \cdots + \theta_{j,k}\exp(x_2)
\end{equation}
\\


Equation~\ref{eq:linearsystem_multid} can be written in a compact form as:

\begin{equation}
    \mathrm{y}_k = \mathrm{U}(\mathrm{x}^T)\mathrm{\theta}_k
    \label{eq:multid}
\end{equation}

The application presented in~\cite{sindyae} uses this approach, where the authors aim to learn differential equations that govern the dynamics of a given system, such as a nonlinear pendulum and the Lorenz system. The approach successfully learned the exact weights, allowing them to recover the correct governing equations.
\\

\begin{figure}[htp]
    \centering
    \begin{tikzpicture}[circed/.style ={circle,fill}]]
        \draw [black, thick] (0.3,0) to [square left brace ] (0.3,2);    
        \draw [line width=2mm, blue] (0.5,0) -- (0.5,2);
        \draw [line width=2mm, red] (0.9,0) -- (0.9,2);
        \draw [line width=2mm, brown] (1.3,0) -- (1.3,2);
        \node at (0.5, 2.2) {$\mathrm{y}_1$};
        \node at (0.9, 2.2) {$\mathrm{y}_2$};
        \node at (1.3, 2.2) {$\mathrm{y}_3$};
        \draw [black, thick] (1.5,0) to [square right brace] (1.5,2);
        \node at (2, 1) {$=$};
        \draw [black, thick] (2.4,0) to [square left brace ] (2.4,2);
        \draw [line width=2mm, gray ] (2.65,0.05) -- (2.65,1.95); \node at (2.65, 2.2) {$1$};
        \draw [line width=2mm, gray ] (3.05,0.05) -- (3.05,1.95); \node at (3.05, 2.2) {$x_1$};
        \draw [line width=2mm, gray ] (3.45,0.05) -- (3.45,1.95); \node at (3.45, 2.2) {$x_2$};
        \draw [line width=2mm, gray ] (3.85,0.05) -- (3.85,1.95); \node at (3.85, 2.2) {$x_3$};
        \draw [line width=2mm, gray ] (4.25,0.05) -- (4.25,1.95); \node at (4.25, 2.25) {\small $x_1^2$};
        \node at (4.7, 1) {$\mathbf{\cdots}$};
        \draw [black, thick] (5,0) to [square right brace] (5,2);
        \draw [black, thick] (5.3,0) to [square left brace ] (5.3,2);
        \draw [line width=2mm, gray ] (5.5,0.05) -- (5.5,1.95); \node at (5.55, 2.2) {$\theta_1$};
        \draw [line width=2mm, gray ] (5.9,0.05) -- (5.9,1.95); \node at (5.95, 2.2) {$\theta_2$};
        \draw [line width=2mm, gray ] (6.3,0.05) -- (6.3,1.95); \node at (6.35, 2.2) {$\theta_3$};
        \draw [black, thick] (6.5,0) to [square right brace] (6.5,2);
        \node [circed,fill=blue, inner sep =1.5 pt] at (5.5, 1.65){};
        \node [circed,fill=blue, inner sep =1.5 pt] at (5.5, 1.45){};
        \node [circed,fill=red, inner sep =1.5 pt] at (5.9, 1.65){};
        \node [circed,fill=red, inner sep =1.5 pt] at (5.9, 1.45){};
        \node [circed,fill=red, inner sep =1.5 pt] at (5.9, 1.25){};
        \node [circed,fill=brown, inner sep =1.5 pt] at (6.3, 1.25){};
        \node [circed,fill=brown, inner sep =1.5 pt] at (6.3, 0.5){};
    \end{tikzpicture}
    \caption{Schematic of the system of Eq.~\ref{eq:yutheta} for the Lorenz system defined by $y_1 = \sigma(x_2-x_1)$, $y_2 = x_1(\rho - x_3) -x_2$, $y_3 = x_1x_2 - \beta x_3$. A library $\mathrm{U}(\mathrm{X})$ of nonlinear functions of the input is constructed. The marked entries in the $\theta$s vectors denote the non-zero coefficients determining which library functions are active for each of the three variables $\{y_1,y_2,y_3\}$.}
    \label{fig:schematicmd}
\end{figure}
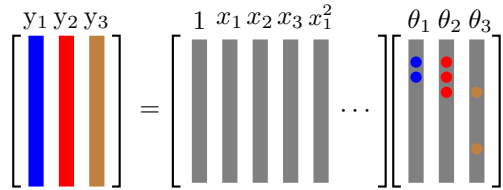

An exemplary schematic is illustrated in Figure~\ref{fig:schematicmd} for the Lorenz system defined by $\dot{x} = \sigma(y-x)$, $\dot{y} = x(\rho - z) -y$, $\dot{z} = xy - \beta z$. Here $x$, $y$, and $z$ are physical variables and $\dot{x}$, $\dot{y}$, and $\dot{z}$ are their respective time-derivatives. Only coefficients associated with functions $\{x_1, x_1x_2,\}$ should be non-zero and equal to the factors shown in the Lorenz system's set of equations.\\

In summary, the linear approach is only successful in particular cases and can not be generalized.  Its main limitation is in predefining the model's structure as a linear combination of nonlinear functions, reducing the SR problem to solve a system of linear equations. In contrast, the main mission of SR is to learn the model's structure and  parameters. A direct consequence of this limitation is that the linear approach fails to learn expressions in many cases: (i) composition of  functions (e.g., $f(x)=f_1(x)*f_2(x)$); (ii) multivariate functions (e.g., $\exp(x*y), \tan(x+y)$, etc.); and (iii) functions including multiplicative or additive factors to their arguments (e.g., $\exp(\lambda x)$). Finally 
the dimension of the library matrix can be challenging in computing resources for extended libraries and high-dimensional data sets.

%% file: nonlinearsr.tex
\section{Nonlinear symbolic regression}
\label{sec:nonlinear}

The nonlinear method uses deep neural networks (DNN), known for their great ability to detect and learn complex patterns directly from data. \\



\begin{figure}[h]
    \centering
        \begin{subfigure}[b]{0.47\textwidth}
        \centering
        \includegraphics[width=.95\linewidth]{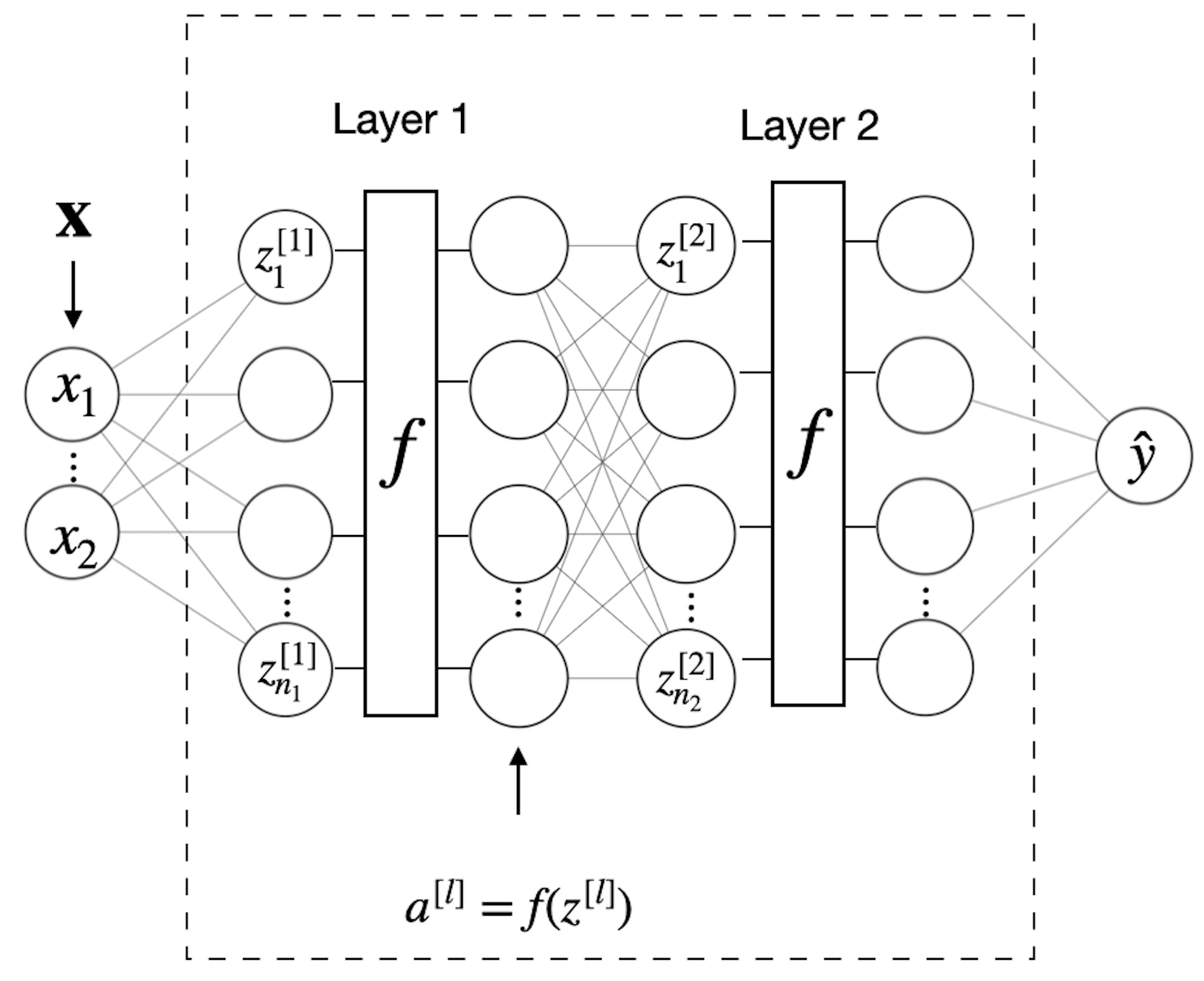}
        \caption{}
        \label{subfig:std}
        \end{subfigure}
        \hfill
        \begin{subfigure}[b]{0.47\textwidth}
        \centering
        \includegraphics[width=.95\linewidth]{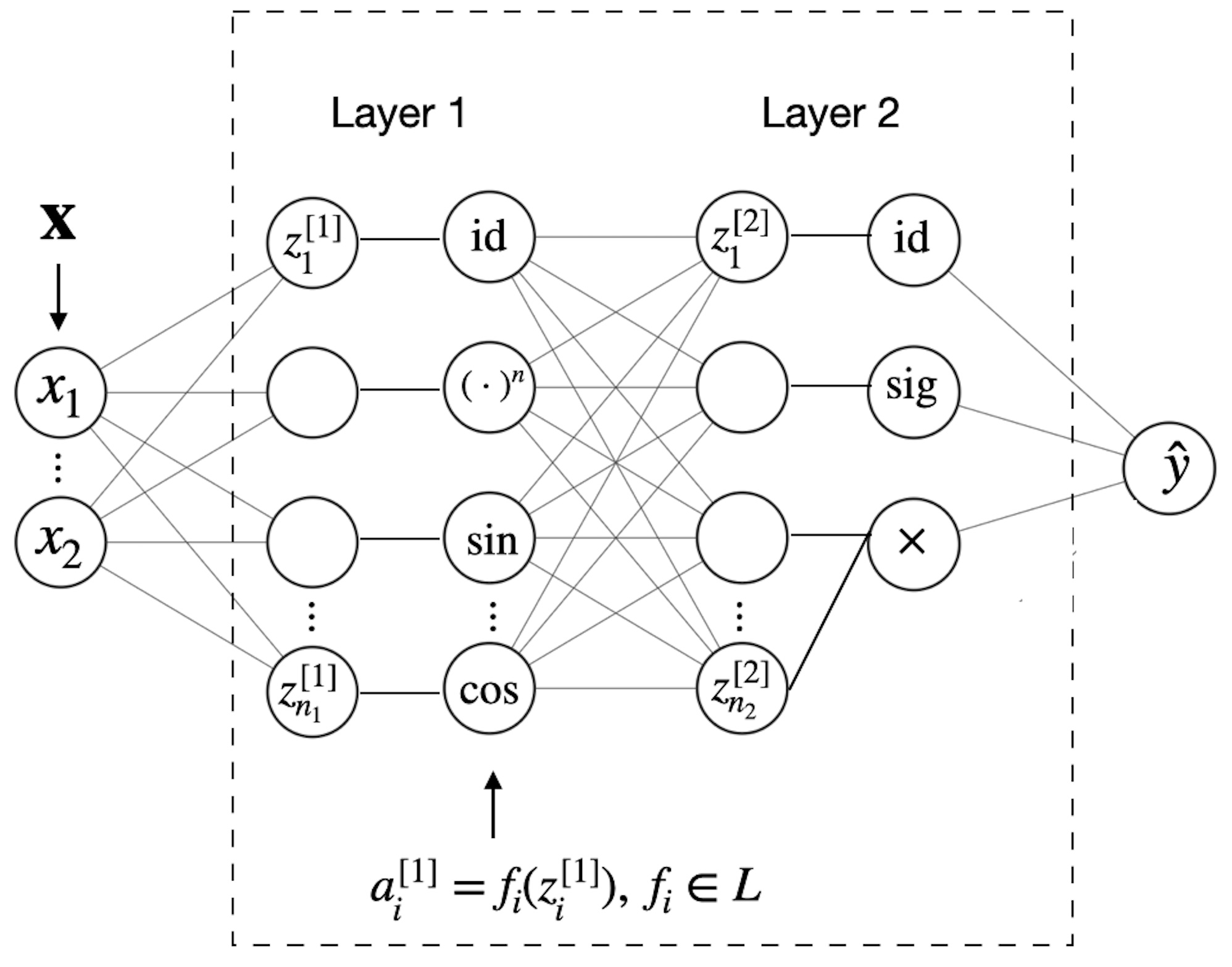}
        \caption{}
        \label{subfig:eql}
        \end{subfigure}
        \hfill
    \caption{Exemplary setup of a standard NN~(\ref{subfig:std}) and  EQL-NN~(\ref{subfig:eql}) with input $\mathbf{x}$, output $\hat{y}$ and two hidden layers. In (a), $f$ denotes the activation function usually chosen among \{RELU,~tanh, sigmoid\} while in EQL each node has a specific activation function drawn from the function class $\mathcal{F}$.}
    \label{fig:NNforSR}
\end{figure}

DNN has the advantage of being fully differentiable in its free parameters allowing end-to-end training using back-propagation. This approach searches the target expression by replacing the standard activation functions in a neural network with elementary mathematical operations. Figure~\ref{fig:NNforSR} shows an NN-based architecture for SR called the Equation Learner (EQL) network proposed by Martius and Lampert~\cite{eql} in comparison with a standard NN. Only two hidden layers are shown for simple visualization, but the network's deepness is controlled as per the case study.\\


The EQL network uses a multi-layer feed-forward NN with one output node. A linear transformation $z^{[l]}$ is applied at every hidden layer ($l$), followed by a nonlinear transformation $a_{i}^{[l]}$ using unary (i.e., one argument) and binary (i.e., two arguments) activation functions as follows

\begin{equation}
    \begin{split}
    z^{[l]} & = W^{[l]}\cdot a^{[l-1]} + b^{[l]}\\
    a^{[l]}_{i} & = f_{i}(z_i^{[l]})
    \end{split}
\end{equation}

where $\{W,b\}$ denote the weight parameters and $f_i$ denotes individual activation function from the library $L = \{\rm{identity},~(\cdot)^{n},~\cos,~\sin,~\exp,~\log,\rm{sigmoid}\}$. In a standard NN, the same activation function is applied to all hidden units and is typically chosen among \{RELU,~tanh,~sigmoid,~softmax, etc.\}.
\\

The problem reduces to learn the correct weight parameters $\{W^{[l]}, b^{[l]}\}$, whereas the operators of the target mathematical expression are selected during training.
To overcome the interpretability limitation of neural network-based architectures and to promote simple over complex solutions as a typical formula describing a physical process, sparsity is enforced by adding a regularization term $l_1$ to the $l_2$ loss function such that,

\begin{equation}
    \ell = \frac{1}{N}\sum_{i=1}^{N}\|\hat{y}(x_i) - y_i \|^2 + \lambda\sum_{l=1}^{L} |W^{[l]}|_{1}
\end{equation}

Where $N$ denotes the number of data entries and $L$ denotes the number of layers. 
Whereas this method is end-to-end differentiable in NN parameters and scales well to high dimensional problems, 
back-propagation through activation functions such as division or logarithm requires simplifications to the search space, thus limiting its ability to produce simple expressions involving divisions (e.g., $\frac{\sin{(x/y)}}{x}$). An extended version EQL$^{\div}$~\cite{eqldiv} includes only the division, whereas exponential and logarithm activation functions are not included because of numerical issues.

%% file: tree.tex
\section{Tree expression}
\label{sec:tree_expression}

\begin{figure}[h]%
\centering
    \begin{subfigure}[b]{0.4\textwidth}
    \centering
        \begin{forest}
        for tree = {circle, draw,
                minimum size=2.em,
                inner sep=0.5pt,
                font=\large,
                l sep=4mm,s sep=3mm
                }
        [$-$[$\times$[$x$][$x$]][\small$\cos$[$x$]]]
        \end{forest}%
        \caption{}
    \label{fig:tree_exp_1}
    \end{subfigure}
    \hfill
     \begin{subfigure}[b]{0.55\textwidth}
         \centering
         \includegraphics[width=.85\linewidth]{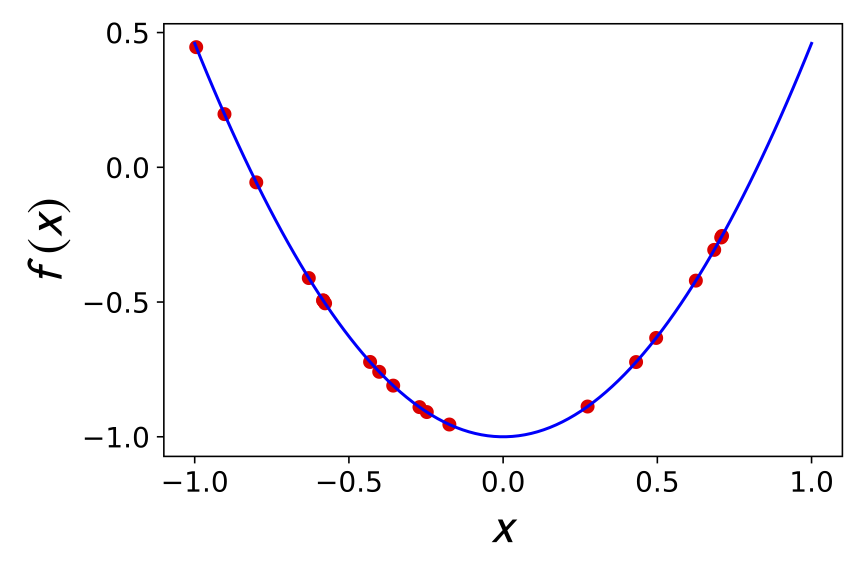}
         \label{fig:tree_exp_2}
         \caption{}
     \end{subfigure}
    \caption{(a) Expression-tree structure of $f(x) = x^2 - \cos(x)$. (b) $f(x)$ as a function of $x$ (blue curve) and data points (red points) generated using $f(x)$.}
\label{fig:treestructure}
\end{figure}

This section discusses SR methods in which a mathematical expression is regarded as a unary-binary tree consisting of internal nodes and terminals. Every tree node represents a mathematical operation (e.g., $+, -, \times, \sin, \log$, etc.) that is drawn from a pre-defined function class $\mathcal{F}$ (Section~\ref{subsec:functionclass}) and every tree terminal node (or leaf) represents an operand, i.e., variable or constant, as illustrated for the example shown in Figure~\ref{fig:treestructure}. Expression tree-based methods include genetic programming, transformers, and reinforcement learning.

\subsection{Genetic Programming}
\label{subsec:gp}

\begin{figure}[h]
    \centering
    \includegraphics[width=.9\linewidth]{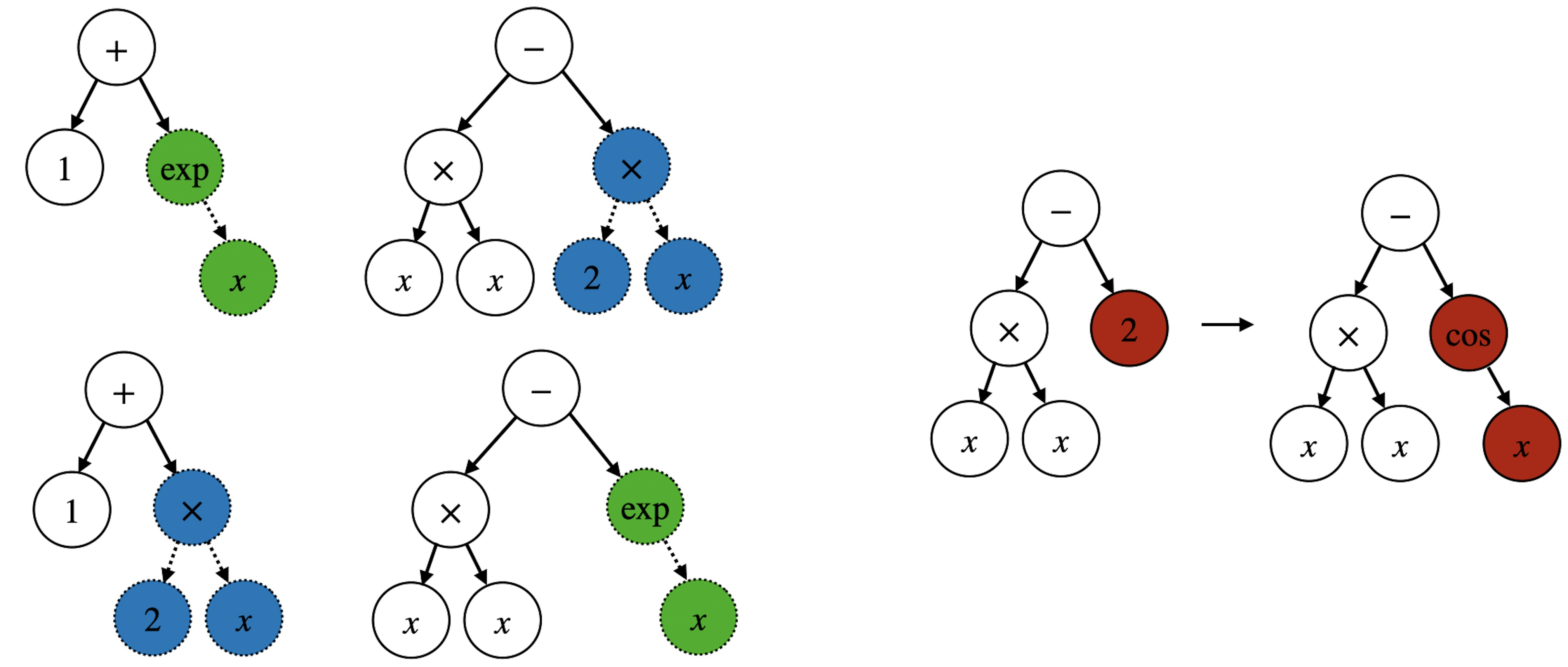}
    \caption{Crossover (left) and mutation (right) operations on exemplary expression trees in genetic programming.}
    \label{fig:gpoperations}
\end{figure}

Genetic programming (GP) is an evolutionary algorithm in computer science that searches the space of computer programs to solve a given problem. Starting with a ``population" (set) of ``individuals" (trees) that is randomly generated, GP evolves the initial population $\mathcal{T}_{GP}^{(0)}$ using a set of evolutionary ``transition rules" (operations) $\{r_i: f\rightarrow f~|~i\in\mathbb{N}\}$ that is defined over the tree space. 
GP evolutionary operations include mutation, crossover, and selection. The mutation operation introduces random variations to an individual by replacing one subtree with another randomly generated subtree (Figure~\ref{fig:gpoperations}-right). The crossover operation involves exchanging content between two individuals, for example, by swapping one random subtree of one individual with another random subtree of another individual (Figure~\ref{fig:gpoperations}-left). Finally, the selection operation is used to select which individuals from the current population persist onto the next population. A common selection operator is tournament selection
, in which a set of $k$ candidate individuals are randomly sampled from the population, and the individual with the highest fitness i.e., a minimum loss is selected. 
%
In a GP algorithm, a single iteration corresponds to one generation. The application of one generation of GP on a population $\mathcal{T}_{GP}^{(i)}$ produces a new, augmented population $\mathcal{T}_{GP}^{(i+1)}$. In each generation, each individual has a probability of undergoing a mutation operation and a probability of undergoing a crossover operation. The selection is applied when the dimension of the current population is the same as the previous one.
Throughout $M_{k}$ iterations, the following steps are undertaken: (1) transition rules are applied to the function set $F^{k}=\{f_1^k,\cdots,f_{M_{k}}^{k}\}$ such that $f^{k+1}=r_i(f^{k})$ where $k$ denotes the iteration index; (2) the loss function $\ell(F^{k})$ is evaluated for the set; and (3) an elite set of individuals is selected for the next iteration step. The GP algorithm repeats this procedure until a pre-determined accuracy level is achieved.\\
 
Whereas GP allows for large variations in the population resulting in improved performance for out-of-distribution data,
GP-based methods do not scale well to high dimensional data sets and are highly sensitive to hyperparameters~\cite{dsr}.

\subsection{Transformers}
\label{subsec:transformers}

Transformer neural network (TNN) is a novel NN architecture introduced by Vaswani \textit{et al.}~\cite{DBLP:journals/corr/VaswaniSPUJGKP17} 
in natural language processing (NLP) to model sequential data. TNN is based on the attention mechanism that aims to model long-range dependencies in a sequence. 
Consider the English-to-French translation of the two following sentences:\\

En: The kid did not go to school because {\bf \underline{it}} was \underline{closed}.\\
Fr: L'enfant n'est pas all\'e \`a l'\'ecole parce qu'\underline{elle} \'etait ferm\'ee.\\
\\
En: The kid did not go to school because {\bf \underline{it}} was \underline{cold}.\\
Fr: L'enfant n'est pas all\'e \`a l'\'ecole parce qu'\underline{il} faisait froid.\\
\\ 

The two sentences are identical except for the last word, which refers to the school in the first sentence (i.e., ``{\bf closed}") and to the weather in the second one (i.e., ``{\bf cold}"). Transformers create a context-dependent word embedding that it pays particular attention to the terms (of the sequence) with high weights. In this example, the noun that the adjective of each sentence refers to has a significant weight and is therefore considered for translating the word ``it". Technically, an embedding $x_i$ is assigned to each element of the input sequence, and a set of $m$ key-value pairs is defined, i.e., $\mathcal{S}=\{(k_1,v_1),\cdots,(k_m,v_m)\}$. For each query, the attention mechanism computes a linear combination of values $\sum_j \omega_jv_j$, where the attention weights ($\omega_j \propto q\cdot k_j$) are derived using the dot product between the query ($q$) and all keys ($k_j$), as follows:

\begin{equation}\label{eq:attention}
\mathrm{Attention}(q,\mathcal{S}) = \sum_j \sigma(q\cdot k_j)v_j
\end{equation}

\begin{figure}[htp]
    \centering
    \includegraphics[width=.85\linewidth]{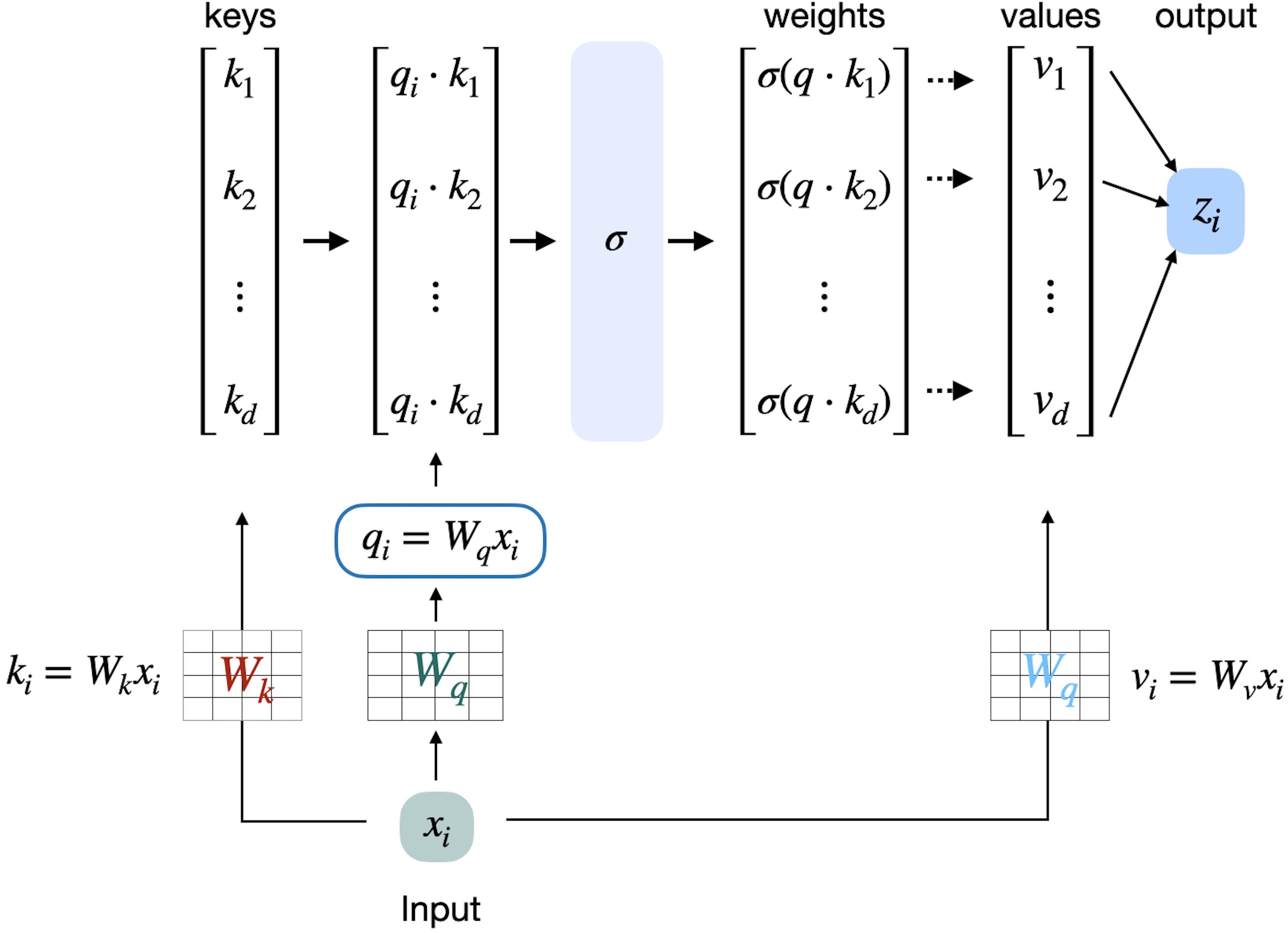}
    \caption{Evaluation of Attention($q,\mathcal{S}$) (Eq.~\ref{eq:attention}) for a query $q_i$, computed using the input vector embedding $x_i$.}
    \label{fig:attention}
\end{figure}

Here, $q=xW_q$ is a query, $k_i = x_iW_{k}$ is a key, $v_i = x_iW_{v}$ is a value, and $W_q$, $W_k$, $W_v$ are the learnable parameters. The architecture of the self-attention mechanism is illustrated in Figure~\ref{fig:attention}.\\

In the context of SR, both input data points $\{(\mathbf{x}_i,y_i)~|~ \mathbf{x}_i\in\mathbb{R}^{d},y_i\in\mathbb{R},~ i\in\mathbb{N}_{n}\}$ and mathematical expressions $f$ are encoded as sequences of symbolic representations as discussed in Section~\ref{subsec:functionrepresentation}.
The role of the transformer is to create the dependencies at two levels, first between numerical and symbolic sequences and between tokens of symbolic sequence. Consider the mathematical expression $f(x,y,z)=\sin(x/y)-\sin(z)$, which can be written as a sequence of tokens following the polish notation:


\begin{table*}[h]
\centering
\begin{tabular}{|c|c|c|c|c|c|c|}\hline
$-$ & $\sin$ & $\div$ & $x$ & $y$ & $\sin$ & $z$\\ \hline
\end{tabular}
\end{table*}

Each symbol is associated with an embedding such that:

\[x_1:-\quad  x_2:\sin\quad
x_3:\div\quad x_4:x\quad 
 x_5:y\quad x_6:\sin\quad x_7:z\]
In this particular example, for query ($x_7:z$), the attention mechanism will give a higher weight for the binary operator ($x_1:-$) than for the variable ($x_5:y$) or the division operator ($x_3:\div$).\\

\begin{figure}[htp]
    \centering
    \includegraphics[width=.5\linewidth]{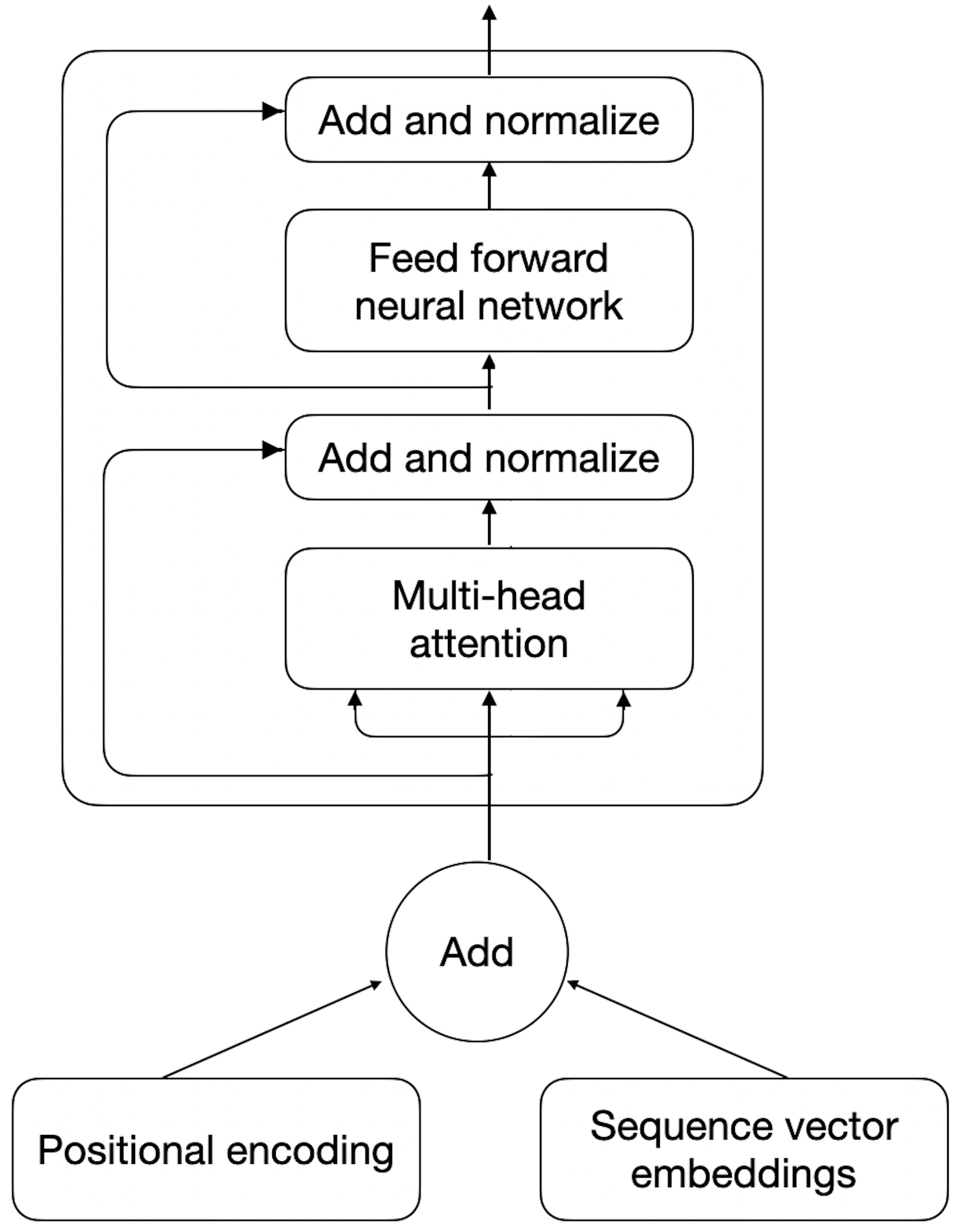}
    \caption{Structure of a TNN encoder~\cite{DBLP:journals/corr/VaswaniSPUJGKP17}. It comprises an attention layer and a feed-forward neural network.}
    \label{fig:encoder}
\end{figure}

Transformers consist of an encoder-decoder structure; each block comprises a self-attention layer and a feed-forward neural network. TNN inputs a sequence of embeddings $\{x_i\}$ and outputs a “context-dependent” sequence of embeddings $\{y_i\}$ one at a time, through a latent representation $z_i$. TNN is an auto-regressive model, i.e., sampling each symbol is conditioned by the previously sampled symbols and the latent sequence. An example of a TNN encoder is shown in Figure~\ref{fig:encoder}.\\

In symbolic regression case, the encoder and the decoder do not share the same vocabulary because the decoder has a mixture of symbolic and numeric representations, while the encoder has only numeric representations.
There exist two approaches to solving SR problems using transformers. First is the skeleton approach~\cite{srthatscales,symbolicgpt} where the transformer conducts the two-steps procedure: (1) the decoder predicts a skeleton $f_e$, a parametric function that defines the general shape of the target expression up to a choice of constants,  using the function class $\mathcal{F}$ and (2) the constants are fitted using optimization techniques such as the non-linear optimization solver BFGS. For example, if $f= \cos(2x_1) -0.1\exp(x_2)$, then the decoder predicts $f_e = \cos(\circ~x_1) -\circ\exp(x_2)$ where $\circ$ denotes an unknown constant. The second is an end-to-end (E2E) approach~\cite{transformers} where both the skeleton and the numerical values of the constants are simultaneously predicted. Both approaches are further discussed in Section~\ref{sec:srapplications}.

\subsection{Reinforcement learning}
\label{subsec:rl}

\begin{figure}[htp]
    \centering
    \includegraphics[width=0.8\linewidth]{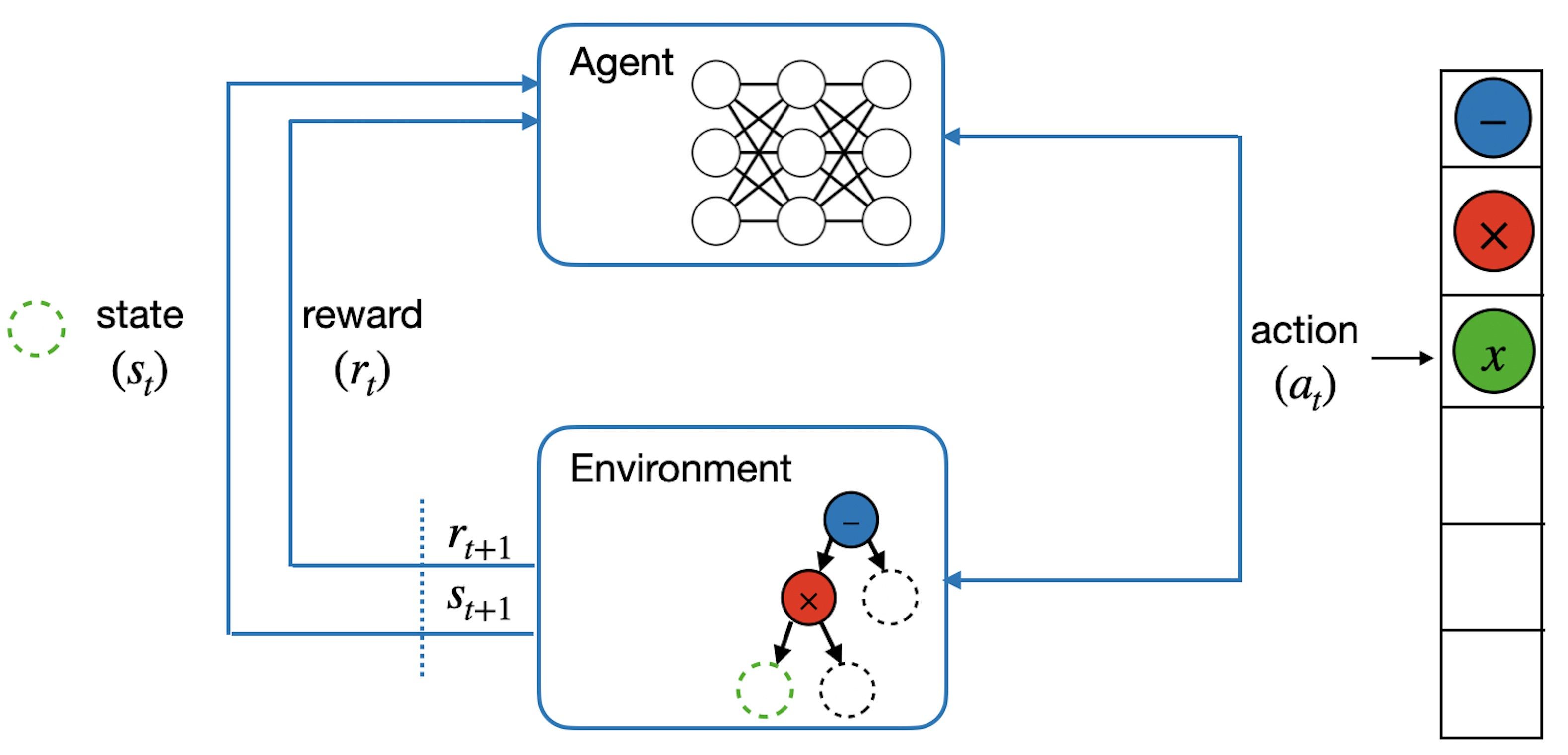}
    \caption{Exemplary sketch of a general RL-based SR method. $s_t$, $a_t$, and $r_t=\mathcal{R}(s_t,a_t)$ denote the state, action, and reward at time step $t$. $(t+1)$ denotes the next time step.}
    \label{fig:rl}
\end{figure}

Reinforcement learning provides a framework for learning and decision-making by trial and error~\cite{sutton_rl}. An RL Setting consists of four components ($\mathcal{S}, \mathcal{A}, \mathcal{P}, \mathcal{R}$) in a Markov decision process. 
In this setting, an agent observes a state $s \in\mathcal{S}$ of the environment and, based
on that, takes action $a \in\mathcal{A}$, which results in a reward $r=\mathcal{R}(s,a)$, and the environment then transitions to a new state $s'\in\mathcal{S}$. The interaction goes
on in time steps until a terminal state is reached. The aim of the agent is to learn the policy $\mathcal{P}$ (also called transition dynamics), which is a mapping from states to actions that maximize the expected cumulative reward. 
An exemplary sketch of an RL-based SR method is illustrated in Figure~\ref{fig:rl}.

SR problem can be framed in RL as follows: the agent (NN) observes the environment (parent and sibling in a tree) and,
based on the observation, takes an action (predict the next token of the sequence) and transitions into a new state. In this view, the NN model is like a policy, the parent and sibling are like observations, and sampled symbols are like actions. 

%% file: srapplications.tex
\section{Applications}\label{sec:srapplications}

\begin{figure}[h]
    \centering
    \includegraphics[width=.85\linewidth]{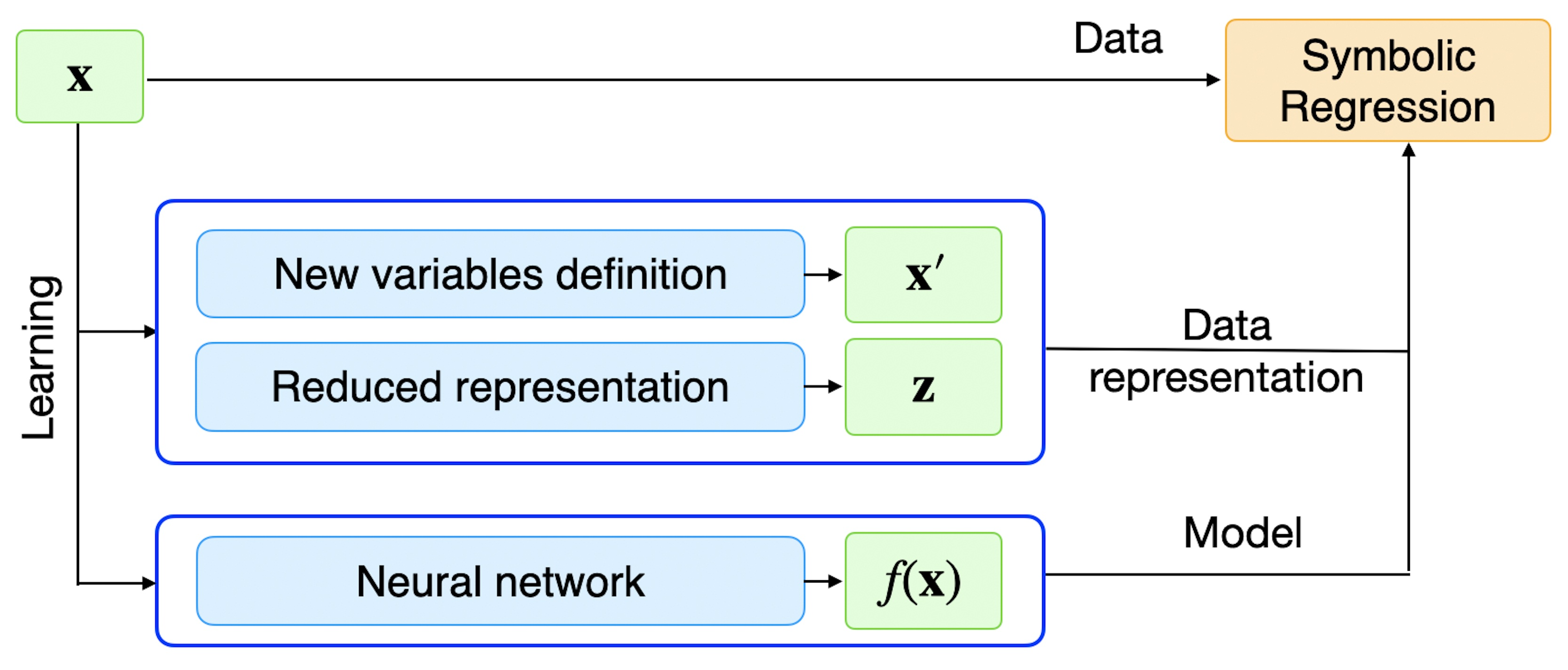}
    \caption{Strategies for solving SR problem. An SR algorithm has three types of input: data ($\mathrm{x}$), a new or reduced representation of the data ($\mathrm{x}^{\prime}$ or $\mathrm{z}$), or a model ($f(\mathrm{x})$) learned from the data.}
    \label{fig:taxonomy_srapplications}
\end{figure}

Most existing algorithms for solving SR are GP-based, whereas many others, and more recent, are deep learning (DL)-based. There exist two different strategies to solve SR problems, as illustrated in the taxonomy of Figure~\ref{fig:taxonomy_srapplications}. 

The first is a one-step approach, where data points are directly fed into an SR algorithm. A second is a two-step approach involving a process which either learns a new representation of data or learns a {\em ``blackbox"} model, which will be then fed into SR algorithm as described below: 

\begin{enumerate}
    \item Learn a new representation of the original data set by defining new features (reducing the number of independent variables) or a reduced representation using specific NN architectures such as principal component analysis and autoencoders.
    \item Learn a {\em ``blackbox"} model either using regular NN or using conceptual NN such as graph neural network (GNN). In this case, an SR algorithm is applied to the learned model or parts of it.
\end{enumerate}

We group the applications based on the categories presented in Section~\ref{sec:srmethods}, and we summarize them in Table~\ref{tab:srapplications}.\\


\begin{table}[htp]
\caption{Table summarizing symbolic regression applications. D refers to data input, 1 refers to data representation input, and 2 refers to model input to SR.}
\label{tab:srapplications}
\centering
\begin{tabular}{llp{8cm}lll}
\toprule
Application & Ref &  Name/description(code) & Year & Method & Strategy\\ 
\midrule
    SINDY & \cite{sindy} & Sparse Identification of Nonlinear Dynamics \href{faculty.washington.edu/sbrunton/sparsedynamics.zip}{link} & 2016 & Linear & D\\
    \\[5pt]
    SINDY-AE & \cite{sindyae} & Data-driven discovery of coordinates and governing equations  (\url{https://github.com/kpchamp/SindyAutoencoders}) & 2019 & Linear  & 1 
    \\[5pt]
    \midrule
    EQL & \cite{eql} & Equation learner (\url{https://github.com/KristofPusztai/EQL}) & 2016 & non linear  & D
    \\[20pt]
    EQL$\div$ & \cite{eqldiv} & Equation learner division (\url{https://github.com/martius-lab/EQL}) & 2018 & Non linear  & D
    \\[5pt]%
    \midrule
    Eureqa & \cite{Dubcakova:2011:GPEM} & Commercial software & 2011 & GP & D 
    \\[5pt]
    FFX & \cite{ffx} & Fast function extraction (\url{https://github.com/natekupp/ffx/tree/master/ffx}) & 2011 & GP  & D
    \\[20pt] 
    ITEA & \cite{ITEA} & Interaction-Transformation Evolutionary Algorithm for SR (\url{https://github.com/folivetti/ITEA/}) & 2019 &   GP  & D 
    \\[20pt] 
    MRGP & \cite{mrgp} & Multiple Regression GP (\url{https://github.com/flexgp/gp-learners}) & 2014 & GP & D  
    \\[5pt] 
    \\ \hdashline \\
    E2ESR & \cite{transformers} & End-to-end SR with transformers (\url{https://github.com/facebookresearch/symbolicregression}) & 2022 & TNN & D  
    \\[20pt]
    NeSymReS & \cite{srthatscales} & Neural SR that scales (\url{https://github.com/SymposiumOrganization/ NeuralSymbolicRegressionThatScales}) & 2021 & TNN & D 
    \\[5pt]
    \\ \hdashline \\
    DSR & \cite{dsr} & Deep symbolic regression (\url{https://github.com/brendenpetersen/deep-symbolic-regression}) & 2019 & RNN,RL & D 
    \\[20pt]
    NGPPS & \cite{dsrgp} & SR via Neural-Guided GP population seeding (\url{https://github.com/brendenpetersen/deep-symbolic-regression}) & 2021 & RNN,GP,RL & D 
    \\[10pt]
    \midrule
    AIFeynman & \cite{Udrescu:2019mnk} & Physics-inspired method for SR (\url{https://github.com/SJ001/AI-Feynman}) & 2019 & Physics-informed & 1 
    \\[20pt]
    \midrule
    SM & \cite{metamodel} & Symbolic Metamodel (\url{https://bitbucket.org/mvdschaar/mlforhealthlabpub}) & 2019 & Mathematics & 2
    \\[20pt]
    \midrule
    GNN & \cite{gnn} & Discovering Symbolic Models from DL with Inductive Biases (\url{https://github.com/MilesCranmer/symbolic_deep_learning}) &  2020 & GNN  & 2
    \\[20pt]
    \midrule
    HEAL & \cite{heal} & Heuristic and Evolutionary Algorithms Laboratory (\url{https://github.com/heal-research/HeuristicLab}) & $-$ & Heuristic & D\\
    \botrule
\end{tabular}
\end{table}

GP-based applications will not be reviewed here; they are listed in the living review~\cite{livingreview}, along with DL-based applications. State-of-the-art GP-based methods are discussed in detail in~\cite{lacava}. Among GP-based applications is the commercial software Eureqa~\cite{Dubcakova:2011:GPEM}, the most well-known GP-based method that uses the algorithm proposed by Schmidt and Lipson in~\cite{doi:10.1126/science.1165893}. Eureqa is used as a baseline SR method in several research works.\\

\textbf{SINDY-AE}~\cite{sindyae} is a hybrid SR method that combines autoencoder  network~\cite{AE} with linear SR~\cite{sindy}.
The novelty of this approach is in simultaneously learning sparse dynamical models and reduced representations of coordinates that define the model using snapshot data. Given a data set $\mathbf{x}(t)\in \mathbb{R}^{n}$, this method seeks to learn coordinate transformations from original to intrinsic coordinates $\mathbf{z}=\phi(\mathbf{x})$ (encoder) and back via $\mathbf{x} = \psi(\mathbf{z})$ (decoder), along with the dynamical model associated with the set of reduced coordinates $\mathbf{z}(t)\in\mathbb{R}^{d}$ ($d\ll n$):

\begin{equation}
    \frac{d}{dt}\mathbf{z}(t) = \mathbf{g}\left(\mathbf{z}(t)\right)
\end{equation}

through a customized loss function $\mathcal{L}$, defined as a sum of four terms:

\begin{equation}
    \mathcal{L} = \underbrace{\|\mathbf{x}-\psi(\phi(\mathbf{x}))\|_2^2}_{\text{reconstruction error}} ~+~
    \lambda_1 \underbrace{\|\mathbf{\dot{z}} -\mathbf{\dot{z}}_{\text{pred}}\|_2^2}_{\text{encoder loss}} ~+~
    \lambda_2\underbrace{\|\mathbf{\dot{x}} - \mathbf{\dot{x}}_{\text{pred}}\|_2^2}_{\text{decoder loss}} ~+~
    \underbrace{\lambda_3 \|\Theta\|_{1}}_{\text{regularizer loss}}
\end{equation}

Here the derivative of the reduced variables $\mathbf{z}$ are computed using the derivatives of the original variable $\mathbf{x}$, i.e. $\mathbf{\dot{z}}=\mathbf{\nabla}_{\mathbf{x}}\phi(\mathbf{x})\dot{x}$.
Predicted coordinates denoted as $\mathbf{a}_{\text{pred}}$ represent NN outputs and are expressed in terms of coefficient vector $\Theta$ and library matrix $\mathbf{U}(\mathbf{x})$ following Eq.~\ref{eq:multid}, i.e., $\mathbf{z}_{\text{rec}} = \mathbf{U}(\mathbf{z}^{T})\Theta = \mathbf{U}(\phi(\mathbf{x})^{T})\Theta$. The library is specified before training, and the coefficients $\Theta$ are learned with the NN parameters as part of the training procedure. 
\\

A case study is the nonlinear pendulum motion whose dynamics are governed by a second-order differential equation given by $\ddot{x}=-\sin(x)$. The data set is generated as a series of snapshot images from a simulated video of a nonlinear pendulum. After training, the SINDY autoencoder correctly identified
the equation $\ddot{z}=-0.99 \sin z$, which is the dynamical model of a nonlinear pendulum in the reduced representation. This approach is particularly efficient when the dynamical model may be dense in terms of functions
of the original measurement coordinates $\mathbf{x}$.
This method and similar works~\cite{lipson_gopro} make the path to ``Gopro physics" where researchers point a camera on an event and get back an equation capturing the underlying phenomenon using an algorithm.\\

Despite successful applications involving partial differential equations, still, one main limitation of this method is in its linear SR part. For example, a model expressed as $f(\mathrm{x}) = x_1x_2 - 2x_2\exp(-x_3) + \frac{1}{2}\exp(-2x_1x_3)$ is discovered only if each term of this expression is comprised in the library, e.g., $\exp(-2x_1x_2)$. The presence of the exponential function, i.e., $\exp(x)$, is insufficient to discover the second and the third terms.
\\

\begin{figure}[htp]
    \centering
    \includegraphics[width=.55\linewidth]{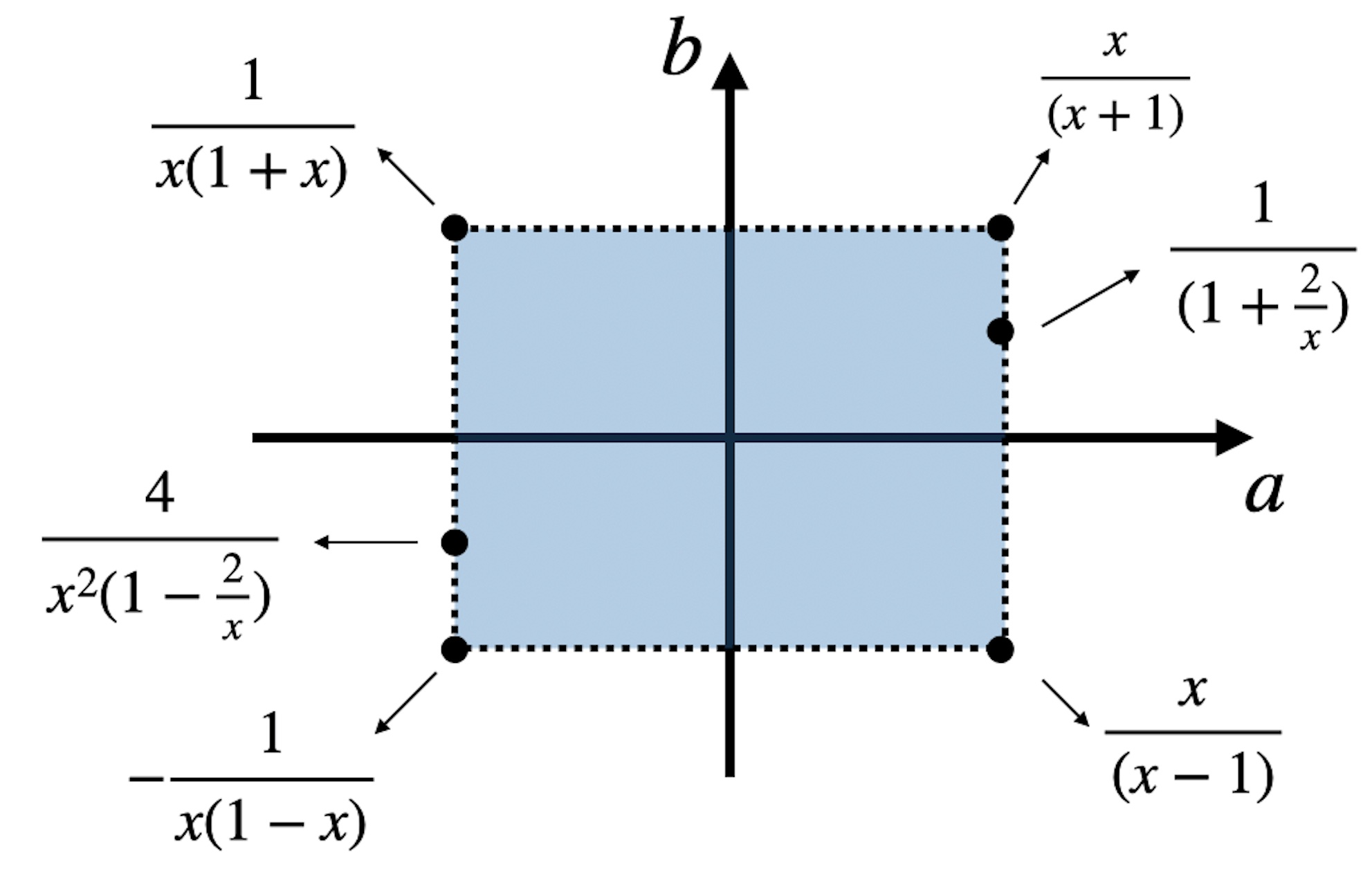}
    \caption{Example of a Meijer G-function $G^{2,2}_{1,1}(^{a,a}_{a,b}|x)$ for different values of $a$ and $b$~\cite{metamodel}.}
    \label{fig:meijer}
\end{figure}

\textbf{Symbolic metamodel}~\cite{metamodel} (SM) is a \textit{model-of-a-model} method for interpreting ``blackbox" model predictions. It inputs a learned {\em ``blackbox"}) model and outputs a symbolic expression. Available post-hoc methods aim to explain ML model predictions, i.e., they can explain some aspects of the prediction but can not offer a full model interpretation. In contrast, SM is interpretable because it uncovers the functional form that underlies the learned model. The symbolic metamodel is based on Meijer $G$-function~\cite{meijer,introtomeijer}, which is a special univariate function characterized by a set of indices, i.e., $G^{m,n}_{p,q}(\mathbf{a}_p,\mathbf{b}_q|x)$, where $\mathbf{a}$ and $\mathbf{b}$ are two sets of real-values parameters. An instance of the Meijer $G$-function is specified by ($\mathbf{a},\mathbf{b}$), for example the function $G^{1,2}_{2,2}(^{a,a}_{a,b}|x)$ takes different forms for different settings of the parameters $a$ and $b$, as illustrated in Figure~\ref{fig:meijer}. \\

In the context of SR problem solving, the target mathematical expression is defined as a parameterization of the Meijer function, i.e., $\{g(x) = G(\theta,\mathbf{x})~|~\theta=(\mathbf{a},\mathbf{b})\}$, thus reducing the optimization task to a standard parameter optimization problem that can be efficiently solved using gradient descent algorithms $\theta^{k+1} := \theta^{k} -\gamma\sum_i l(G(\mathbf{x}_i,\theta),f(\mathbf{x}_i))|_{\theta=\theta^{k}}$. The parameters $\mathbf{a}$ and $\mathbf{b}$ are learned during training, and the indices $(m,n,p,q)$ are regarded as hyperparameters of the model. SM was tested on both synthetic and real data and was deployed in two modes spanning (1) only polynomial expressions (SM$^{p}$) and (2) closed-form expressions (SM$^{c}$), in comparison to a GP-based SR method. SM$^{p}$ produces accurate polynomial expressions for three out of four tested functions (except the Bessel function), whereas SM$^{c}$ produces the correct ground-truth expression for all four functions and significantly outperforms GP-based SR.\\

More generally, consider a problem in a critical discipline such as healthcare. Assuming a feature vector comprising (age, gender, weight, blood pressure, temperature, disease history, profession, etc.) with the aim to predict the risk of a given disease. Predictions made by a {\em ``blackbox"} could be highly accurate. However, the learned model does not provide insights into why the risk is high or low for a patient and what parameter is the most critical or weightful in the prediction. Applying the symbolic metamodel to the learned model outputs a symbolic expression, e.g., $f(x_1,x_2) = x_1\left(1 - \exp(-x_2)\right)$, where $x_1$ is the blood pressure and $x_2$ is the age. Here, we can learn that only two features (out of many others) are crucial for the prediction and that the risk increases with high blood pressure and  decreases with age. This is an ideal example showing the difference between {\em ``blackbox"} and interpretable models. In addition, it is worth mentioning that methods applied for model interpretation only exploit part of the prediction and can not unveil how the model captures nonlinearities in the data. Thus model interpretation methods are insufficient to provide full insights into why and how model predictions are made and are not by any means equivalent to interpretable models.
\\

\textbf{End-to-end symbolic regression}~\cite{transformers} (E2ESR) is a transformer-based method that uses end-to-end learning to solve SR problems. It is made up of three components: (1) an embedder that maps each input point $(x_i,y_i)$ to a single embedding, (2) a fully-connected feedforward network, and (3) a transformer that outputs a mathematical expression. What distinguishes E2ESR from other transformer-based applications is the use of an end-to-end approach without resorting to skeletons, thus using both symbolic representations for the operators and the variables and numeric representations for the constants. Both input data points $\{(\mathbf{x}_i,y_i)~|~ i\in\mathbb{N}_{n}\}$ and mathematical expressions $f$ are encoded as sequences of symbolic representations following the description in Section~\ref{subsec:functionrepresentation}. E2ESR is tested and compared to several GP-based and DL-based applications on SR benchmarks. Results are reported in terms of mean accuracy, formula complexity, and inference time, and it was shown E2ESR achieves very competitive results for SR and outperforms previous applications.
\\

\textbf{AIFeynman}~\cite{Udrescu:2019mnk} is a physics-inspired SR method that recursively applies a set of solvers, i.e., dimensional analysis\footnote{Dimensional analysis is a well-known technique in physics that uses set of units of measurements to solve an equation and/or to check the correctness of a given equation.}, polynomial fit, and brute-force search to solve an SR problem. If the problem is not solved, the algorithm searches for simplifying intrinsic properties in data (e.g. invariance, factorization) using NN and deploys them to recursively simplify the dataset into simpler sub-problems with fewer independent variables. Each sub-problem is then tackled by a symbolic regression method of choice. The authors created the Feynman SR database~(see Section~\ref{sec:data}) to test their approach. All the basic equations and 90\% of the bonus equations were solved by their algorithm, outperforming Eureqa.\\

\begin{figure}[htp]
    \centering
    \includegraphics[width=1.\linewidth]{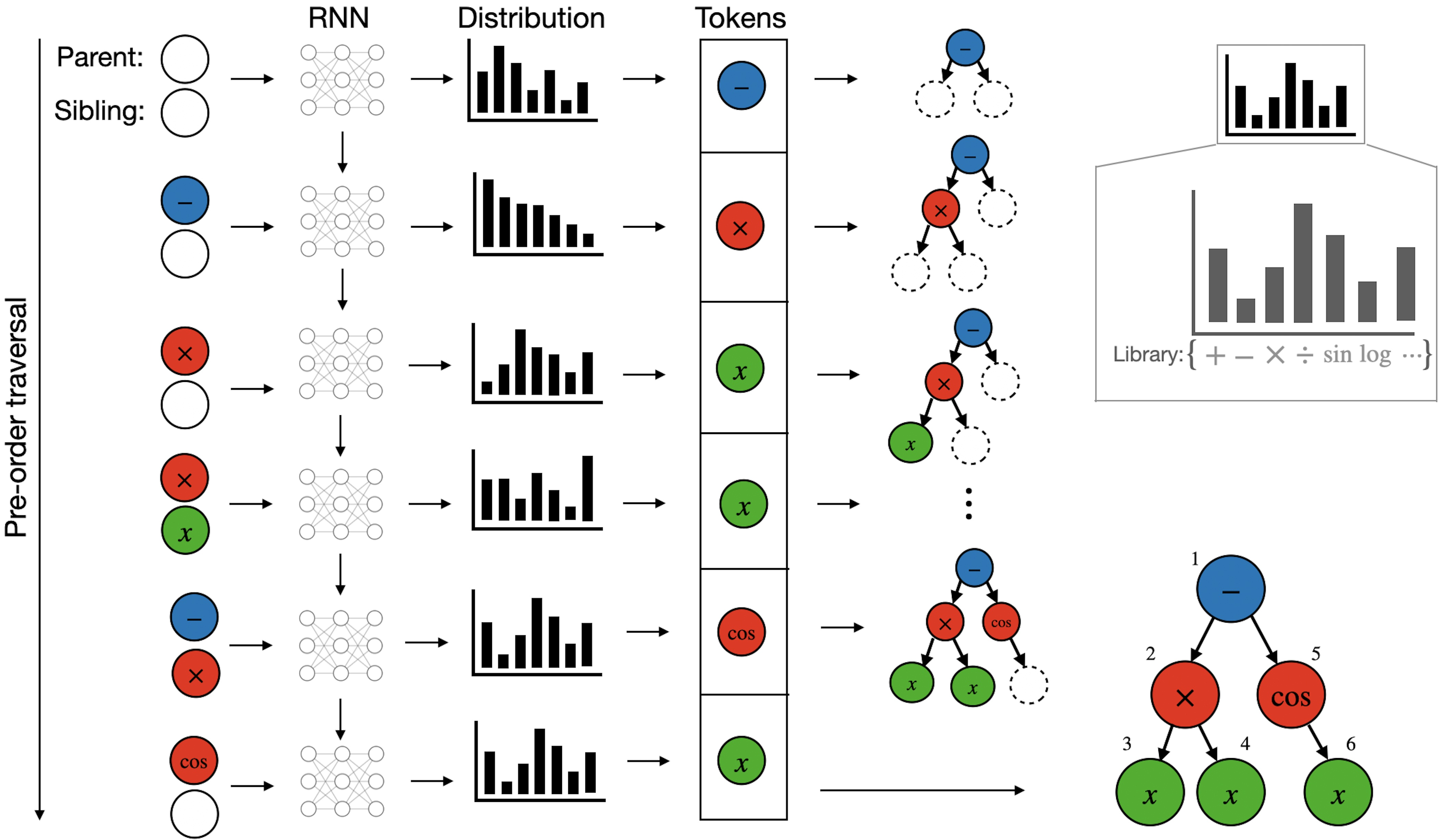}
    \caption{Exemplary sketch of RNN generating a mathematical expression $x^2 - \cos(x)$.}
    \label{fig:dsr_rnn}
\end{figure}

\textbf{Deep Symbolic Regression} (DSR)~\cite{dsr} is an RL-based search method for symbolic regression that uses a generative recurrent neural network (RNN). 
RNN defines a probability distribution ($p(\theta)$) over mathematical expressions ($\tau$), and batches of expressions $\mathcal{T}=\{\tau^{(i)}\}_{i=1}^{N}$ are stochastically generated. 
An exemplary sketch of how RNN generates an expression (e.g., $x^2 - \cos(x)$) is shown in Figure~\ref{fig:dsr_rnn}. Starting with the first node following the pre-order traversal (Section~\ref{subsec:functionrepresentation}) of an expression tree, RNN is initially fed with empty placeholders tokens (a parent and a sibling) and produces a categorical distribution, i.e., outputs the probability of selecting every token from the defined library $L = \{+,-,\times,\div,\sin,\cos,\log,\mathrm{etc}.\}$. The sampled token is fed into the first node, and the number of siblings is determined based on whether the operation is unary (one sibling) or binary (two siblings). The second node is then selected, and the RNN is fed with internal weights along with the first token and outputs a new (and potentially different) categorical distribution. This procedure is repeated until the expression is complete.
Expressions are then evaluated with a reward function $R(\tau)$ to test the goodness of the fit to the data $\mathcal{D}$ for each candidate expression $(f)$ using normalized root-mean-square error,  $R(\tau) = 1/\left(1+\frac{1}{\sigma_y}\sqrt{\frac{1}{n}\sum_{i=1}^{n}(y_i-f(\mbox{X}_i))^2}\right)$. 

To generate better expressions ($f$), the probability distribution $p(\tau|\theta)$ needs to be optimized. Using a gradient-based approach for optimization requires the reward function $R(\tau)$ to be differentiable with respect to the RNN parameter $\theta$, which is not the case. Instead, the learning objective is defined as the expectation of the reward under expressions from the policy, i.e., $J(\theta) = \mathbb{E}_{\tau\sim p(\tau|\theta)}[R(\tau)]$, and reinforcement learning is used to maximize $J(\theta)$ by means of the ``standard policy gradient":

\begin{equation}
\begin{split}
 \nabla_{\theta}J(\theta) 
    = \nabla_{\theta}\mathbb{E}_{\tau\sim p(\tau|\theta)}[R(\tau)]
    = \mathbb{E}_{\tau\sim p(\tau|\theta)}[R(\tau)\nabla_{\theta}\log p(\tau|\theta)]\\
\end{split}
\label{eq:spg}
\end{equation}

This reinforcement learning trick, called REINFORCE~\cite{reinforce}, can be derived using the definition of the expectation $\mathbb{E}[\cdot]$ and the derivative of $\log(\cdot)$ function as follows:

\begin{equation}
\begin{split}
    \nabla_{\theta}\mathbb{E}_{\tau\sim p(\tau|\theta)}[R(\tau)] &= \nabla_{\theta}\int R(\tau)p(\tau|\theta)d\theta\\
    &= \int R(\tau)\nabla_{\theta}p(\tau|\theta)d\theta\\
    &= \int R(\tau)\frac{\nabla_{\theta}p(\tau|\theta)}{p(\tau|\theta)}p(\tau|\theta)d\theta\\
    &= \int R(\tau) \log(p(\tau|\theta)p(\tau|\theta)d\theta\\
    &= \mathbb{E}_{\tau\sim p(\tau|\theta)}[R(\tau)\nabla_{\theta}\log p(\tau|\theta)]\\
\end{split}
\end{equation}

The importance of this result is that it allows estimating the expectation using samples from the distribution. More explicitly, the gradient of $J(\theta)$ is estimated by computing the mean over a batch of N sampled expressions as follows:

\begin{equation}
    \nabla_{\theta}J(\theta) 
    = \frac{1}{N}\sum_{i=1}^{N}R(\tau^{(i)})\nabla_{\theta}\log p(\tau^{(i)}|\theta)
\end{equation}

The standard policy gradient (Eq.~\ref{eq:spg}) permits optimizing a policy's average performance over all samples from the distribution. 
Since SR requires maximizing best-case performance, i.e., to optimize the gradient over the top $\epsilon$ fraction of samples from the distribution found during training, a new learning objective is defined as a conditional expectation of rewards above the $(1-\epsilon)$-quantile of the distribution of rewards, as follows:

\begin{equation}
    J_{\mathrm{risk}}(\theta,\epsilon) = \mathbb{E}_{\tau\sim p(\tau|\theta)}[R(\tau)~|~ R(\tau) \geq R_{\epsilon}(\theta)]
\end{equation}

where $R_{\epsilon}(\theta)$ represent the samples from the distribution below the $\epsilon$-threshold. The gradient of the new learning objective is given by:

\begin{equation}
     \nabla_{\theta}J_{\mathrm{risk}}(\theta) 
    = \mathbb{E}_{\tau\sim p(\tau|\theta)}[(R(\tau)-R_{\epsilon}(\theta))\cdot\nabla_{\theta}\log p(\tau|\theta) ~|~ R(\tau)\geq R_{\epsilon}(\theta)]
\end{equation}

DSR was essentially evaluated on the Nguyen SR benchmark and several additional variants of this benchmark. An excellent recovery rate was reported for each set, and DSR solved all mysteries except the Nguyen-12 benchmark given by $x^4 -x^3 +\frac{1}{2}y^2 -y$. More details on SR data benchmarks can be found in Section~\ref{sec:data}.\\

\textbf{Neural-guided genetic programming population seeding}~\cite{dsrgp} (NGPPS) is a hybrid method that combines GP and RNN~\cite{dsr} and leverages the strengths of each of the two components. Whereas GP begins with random starting populations, the authors in~\cite{dsrgp} propose to use the batch of expressions sampled by RNN as a staring population for GP: $\mathcal{T}_{GP}^{(0)} = \mathcal{T}_{RNN}$. Each iteration of the proposed algorithm consists of 4 steps: (1) The batch of expressions sampled by RNN is passed as a starting population to GP, (2) $S$ generations of GP are performed and result in a final GP population $\mathcal{T}_{GP}^{S}$, (3) An elite set of top-performing GP samples is selected $\mathcal{T}_{GP}^{E}$ and passed to the gradient update of RNN.\\

\begin{figure}[htp]
    \centering
    \includegraphics[width=.65\linewidth]{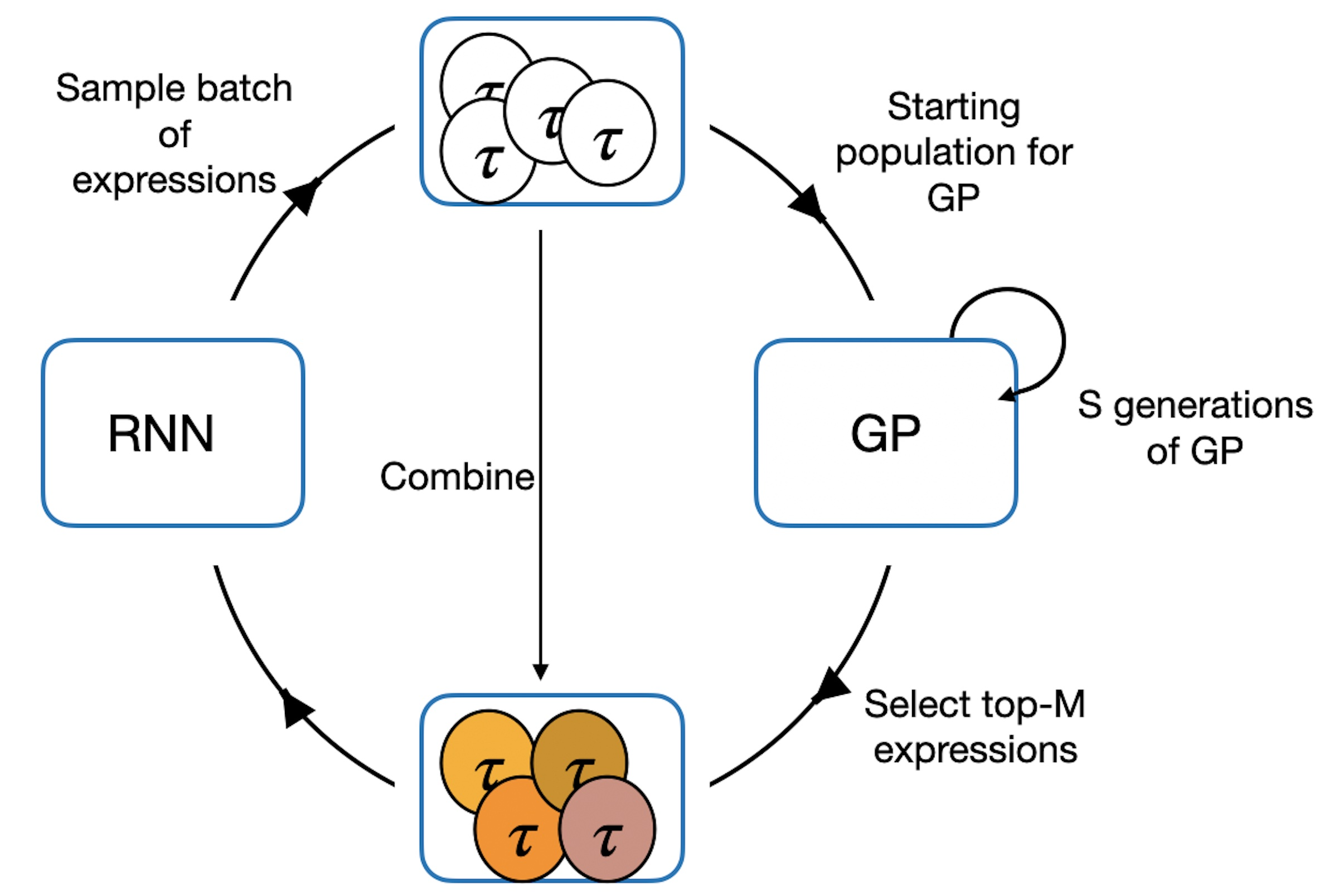}
    \caption{Neural-guided genetic programming population seeding method overview~\cite{dsrgp}.}
    \label{fig:dsrgp}
\end{figure}

\textbf{Neural symbolic regression that scales}~\cite{srthatscales} (NeSymReS) is a transformer-based algorithm that emphasizes large-scale pre-training. It comprises a pre-training and test phase. Pre-training includes data generation and model training. Hundreds of millions of training examples are generated for every minibatch in pre-training. Each training example consists of a symbolic equation $f_e$ and a set of $n$ input-output pairs $\{\mathbf{x}_i, y_i=f(\mathbf{x}_i)\}$ where $n$ can vary across examples, and the number of independent input variables is at most three. In the test phase, a set of input-output pairs $\{x_i,y_i\}$ is fed into the encoder that maps it into a latent vector $z$, and the decoder iteratively samples candidates' skeletons. What distinguishes this method is the learning task, i.e., it improves over time with experience, and there is no need to be retrained from scratch on each new experiment.
It was shown that NeSymReS outperforms selected baselines (including DSR) in time and accuracy by a large margin on all datasets (AI-Feynman, Nguyen, and strictly out-of-sample equations (SOOSE) with and without constants).  
NeSymReS is more than three orders of magnitudes faster at reaching the same maximum accuracy as GP while only running on CPU.\\

\textbf{GNN}~\cite{gnn} is a hybrid scheme performing SR by training a Graph Neural Network (GNN) and applying SR algorithms on GNN components to find mathematical equations.\\

A case study is Newtonian dynamics which describes the dynamics of particles in a system according to Newton's laws of motion. $\mathcal{D}$ consists of an N-body system with known interaction (force law $F$ such as electric, gravitation, spring, etc.), where particles (nodes) are characterized by their attributes (mass, charge, position, velocity, and acceleration) and their interaction (edges) are assigned the attribute of dimension 100. The GNN functions are trained to predict instantaneous acceleration for each particle using the simulated data and then 
applied to a different data sample. The study shows that the most significant edge attributes, say $\{e_1, e_2\}$, fit to a linear combination of the true force components, $\{F_1, F_2\}$, which were used in the simulation showing that edge attributes can be interpreted as force laws.
The most significant edge attributes were then passed into Eureqa to uncover analytical expressions that are equivalent to the simulated force laws. 
The proposed approach was also applied to datasets in the field of cosmology, and it discovered an equation that fits the data better than the existing hand-designed equation.\\

The same group has recently succeeded in inferring Newton's law for gravitational force using GNN and \textit{PySR} for symbolic regression task~\cite{cranmersolar}. GNN was trained using observed trajectories (position) of the Sun, planets, and moons of the solar system collected during 30 years. The SR algorithm could correctly infer Newton's formula that describes the interaction between masses, i.e., $F=-GM_1M_2/r^2$, and the masses and the gravitational constant as well.

%% file: datasets.tex
\section{Datasets}
\label{sec:data}

\begin{figure}[h]
    \centering
    \begin{forest}
    for tree={%
        my node,
        l sep+=5pt,
        grow'=east,
        edge={gray, thick},
        child anchor=north,
        parent anchor=south,
        grow=south,
        align=center,
        edge path={
        \noexpand\path[line width=1pt, \forestoption{edge}]
        (!u.parent anchor)  |- ([yshift=4pt].child anchor) -- (.child anchor) \forestoption{edge label};
        },
        if n children=0{tier=last}{},
        if={isodd(n_children())}{
            for children={
                if={equal(n,(n_children("!u")+1)/2)}{calign with current}{}
            }
        }{}
    }
    [SR Benchmark
    [Ground-truth problems[Physics equations][Mathematics equations]]
    [Real-world problems[Observations][Measurements]]
    ]
    \end{forest}
    \caption{Taxonomy based on the type of SR benchmark problems.}
    \label{fig:taxonomy_data}
\end{figure}
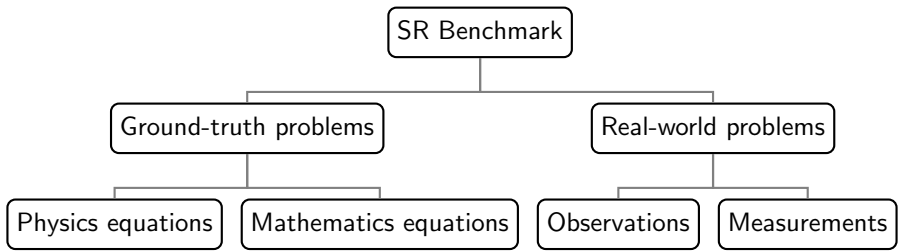

For symbolic regression purposes, there exist several benchmark data sets that can be categorized into two main groups: (1) ground-truth problems (or synthetic data) and (2) real-world problems (or real data), as summarized in Figure~\ref{fig:taxonomy_data}. In this section, we describe each category and discuss its main strength and limitations.\\

\tikzset{
    my node/.style={
        align=center,
        draw=black,
        inner color=white,
        outer color=white,
        thick,
        minimum width=0.9cm,
        rounded corners=3,
        font=\sffamily,
    }
}

Ground-truth regression problems are characterized by known mathematical equations, they are listed in Table~\ref{tab:datasets}. These include (1) physics-inspired equations~\cite{Udrescu:2019mnk,strogatz} and (2) real-valued symbolic equations~\cite{koza,keijer,vladislavleva,Korns2011,Uy2010SemanticallybasedCI,jin,dsr,R}.

\begin{table}[h]
\caption{Table summarizing ground-truth problems for symbolic regression.}
\label{tab:datasets}
\begin{center}
\begin{tabular}{@{}lllll@{}}
\toprule
Type & Benchmark & Number of problems & Year & Reference \\
\midrule
\multirow{2}{*}{Physics-related} & Feynman Database & 119 & 2019 & \cite{Udrescu:2019mnk} \\[5pt]
& Strogatz Repository & 10 & 2011 & \cite{strogatz}\\[5pt] 
\midrule
\multirow{8}{*}{Mathematics-related}  
& Koza & 3 & 1994 & \cite{koza} \\
& Keijzer & 15 & 2003 & \cite{keijer}\\
& Vladislavleva & 8 & 2009 & \cite{vladislavleva}\\
& Nguyen  & 12  & 2011 & \cite{Uy2010SemanticallybasedCI}\\
& Korns & 15 & 2011 & \cite{Korns2011}\\
& R & 3 & 2013 & \cite{R}\\
& Jin & 6 & 2019 &  \cite{jin}\\
& Livermore & 22 & 2021 & \cite{dsr} \\
\botrule
\end{tabular}
\end{center}
\end{table}

The Feynman Symbolic Regression Database~\cite{AIFeynmanSR} is the largest SR database that originates from Feynman lectures on Physics series~\cite{feynman2011feynman,feynman2006feynman} and is proposed in~\cite{Udrescu:2019mnk}. It consists of 119\footnote{The equation number  II.11.17 is missing in the benchmark repository.} physics-inspired equations that describe static physical systems and various physics processes. The proposed equations depend on at least one variable and, at most, nine variables. Each benchmark (corresponding to one equation) is generated by randomly sampling one million entries. Each entry is a row of randomly generated input variables, which are sampled uniformly between 1 and 5. This range of sampling was slightly adjusted for some equations to avoid unphysical results (e.g., division by zero or the square root of a negative number). The output is evaluated using function $f$, e.g. $\mathcal{D}=\{\mathbf{x}_i\in\mathbb{R}^{d}, y
_i=f(x_1,\cdots,x_d)\}$.

This benchmark is rich in proposing various theoretical formulae. Still, it suffers a few limitations: (1) there is no distinction between variables and constants, i.e., constants are randomly sampled and, in some cases, in domains extremely far from physical values. For example, the speed of light is sampled from a uniform distribution $\mathcal{U}(1,20)$ whereas its physical value is orders of magnitude higher, i.e., $c=2.988\times10^8$ m/s, and the gravitational constant is sampled from $\mathcal{U}(1,2)$ whereas its physical value is orders of magnitude smaller, $G = 6.6743\times10^{-11}$ m$^3$ kg$^{-1}$ s$^{-2}$, among others (e.g., vacuum permittivity $\epsilon\sim10^{-12}$, Boltzmann constant $k_{b}\sim10^{-23}$, Planck constant $h\sim10^{-34}$). (2) Some variables are sampled in nonphysical ranges. For example, the gravitational force is defined between two masses distant by $r$ as $F = Gm_1m_2/r^2$. This force is weak unless defined between significantly massive objects (e.g., the mass of the earth is $M_e = 5.9722\times10^{24}$ kg) whereas $m_1$ and $m_2$ are sampled in $\mathcal{U}(1,5)$ in the Feynman database. (3) Some variables are treated as floats while they are integers, and (4) many equations are duplicates of each other (e.g., a multiplicative function of two variables $f(x,y)=x*y$) or have similar functional forms.\\ 
%
The ODE-Strogatz repository~\cite{strogatz} consists of ten physics equations that describe the behavior of dynamical systems which can exhibit chaotic and/or non-linear behavior. Each dataset is one state of a two-state system of ordinary differential equations.\\

Within the same category, there exist several benchmarks~\cite{koza,keijer,vladislavleva,Korns2011,Uy2010SemanticallybasedCI,jin,dsr} consisting of real-valued symbolic functions. The majority of these benchmarks are proposed for GP-based methods and grouped into four categories: polynomial, trigonometric, logarithmic, exponential, and square-root functions, and a combination of univariate and bivariate functions. The suggested functions do not have any physical meaning, and most depend either on one or two independent variables. Datasets are generally generated by randomly sampling either 20 or 100 points in narrow ranges.
The most commonly known is the so-called Nguyen benchmark, which consists of 12 symbolic functions taken from~\cite{keijer2003,Hoai,johnson}. Only four equations have the scalars \{1,2,1/2\} as constants therein.
Each benchmark is defined by a ground-truth expression, training, and test datasets.
The equations proposed in these benchmarks can not be found in a single repository. Therefore we list them in the appendix in Tables~[\ref{tab:srprob1}-\ref{tab:srprob3}] and Tables ~[\ref{tab:aifeynman1}-\ref{tab:aifeynman3}] for completeness and for easy comparison.\\

Real-world problems are characterized by an unknown model that underlies data. 
This category comprises two groups: observations and measurements. 
Data sets in the observations category can originate from any domain, such as health informatics, environmental science, business, commerce, etc. Data could be collected online or offline from reports or studies. A wide range of problems can be assessed from the following repositories: the PMLB~\cite{Olson2017PMLB}, the OpenML~\cite{OpenML2013}, and the UCI~\cite{Dua:2019}. An exemplary application in this category is wind speed forecasting~\cite {abdellaoui2021symbolic}. 
Measurements represent sets of data points that are collected (and sometimes analyzed) in physics experiments.
Here the target model is either an underlying theory than can be derived from first principles or not. In the first case, symbolic regression would either infer the correct model structure and parameters or contribute to the theory development of the studied process, whereas in the second case, the symbolic regression output could be the awaited theory.

%% file: conclusion.tex
\section{Discussion}\label{sec:discussion}

SR is a growing area of ML and is gaining more attention as interpretability is increasingly promoted~\cite{rudin} in AI applications. SR is propelled by the fact that ML models are becoming very big in parameters at the expense of making accurate predictions. An exemplary application is the chatGPT-4, a large language model comprising hundreds of billions of parameters and trained on hundreds of terabytes of textual data. Such big models are very complicated networks. ChatGPT-4, for example, is accomplishing increasingly complicated and intelligent tasks to the point that it is showing emergent properties~\cite{wei2022emergent}. However,  it is not straightforward to understand when it works and, more importantly, when it does not. In addition, its performance improves with increasing the number of parameters, highlighting that its prediction accuracy depends on the size of the training data set. 
Therefore, a new paradigm is needed, especially in scientific disciplines, such as physical sciences, where problems are of causal hypothesis-driven nature. SR is by far the most potential candidate to fulfill the interpretability requirements and is expected to play a central role in the future of ML.
\\

Despite the significant advances made in this subfield and the high performance of most deep learning-based SR methods proposed in the literature, still, SR methods fail to recover
relatively simple relationships. A case in point is the 
Nguyen-12 expression, i.e., $f(x,y) = x^4-x^3+y^2/2 -y$, where $x$ and $y$ are uniformly sampled in the range $[0,1]$. 
The NGPPS method could not recover this particular expression using the 
library basis $L=\{+, -, \times, \div, \sin, \cos, \exp, \log, x, y\}$. A variant of this expression, Nguyen-12$^{\star}$, consisting of the same equation but defined  over a larger domain, i.e., data points sampled in $[0,10]$, was successfully covered using 
the same library, with a recovery rate of $12\%$. This result is significantly below the perfect performance on all other Nguyen expressions. A similar observation is made for the Livermore-5 whose expression is $f(x,y)=x^4-x^3+x^2-y$.
We ran  NGPPS on Nguyen-12 with two libraries, a pure polynomial basis $L_1=\{+, -, \times, \div, (\cdot)^2, (\cdot)^3, (\cdot)^4, x, y\}$ and a mixed basis $L_2=L_1\cup \{\sin,\cos,\exp,\log,\mathrm{sqrt},\mathrm{expneg}\}$. 
The algorithm succeeds in recovering Nguyen-12 only using a pure polynomial basis with a recovery rate of $3\%$. The same observation is made by applying linear SR on Nguyen-12. This highlights how strongly the predicted expression depends on the set of allowable mathematical operations. A practical way to encounter this limitation is to implement basic domain knowledge in SR applications whenever possible. For example, astronomical data collected by detecting the light curves of astronomical objects exhibit periodic behavior. In such cases, periodic functions such as trigonometric functions should be part of the library basis.\\

Most SR methods are only applied to synthetic data for which the input-output relationship is known. This is justified because the methods must be cross-checked, and their performance must be evaluated using ground-truth expressions. However, the reported results are for synthetic data only. To the best of our knowledge, only one physics application~\cite{cranmersolar} succeeded in extracting New's laws of gravitation by applying SR to astronomical data. The absence of such applications leads us to state that
SR is still 
a relatively nascent area with the potential to make a big impact. Physics in  general, and physical sciences in particular, represent a very broad field for SR development purposes and are very rich both in data and expressions, e.g., areas such as astronomy and high-energy physics are very rich in data. In addition, lots of our acquired knowledge in physics can be used for SR methods test purposes because underlying phenomena and equations are well known. All that is needed is greater effort and investment.
\\



\section{Conclusion}\label{sec:conclusion}


This work presents an in-depth introduction to the symbolic regression problem and an expansive review of its methodologies and state-of-the-art applications. Also, this work highlights a number of conclusions that can be made about symbolic regression methods, including (1) linear symbolic regression suffer many limitations, all originating from predefining the model structure, (2) neural network-based methods lead to numerical issues and the library can not include all mathematical operations, (3) expression tree-based methods are yet the most powerful in terms of model performance on synthetic data, in particular transformer-based ones, (4) model predictions strongly depend on the set of allowable operations in the library basis, and (5) generally, deep learning-based methods are performing better than other ML-based methods.
\\

Symbolic regression represents a powerful tool for learning interpretable models in a data-driven manner. Its application is likely to grow in the future because it balances prediction accuracy and interpretability. Despite the limited SR application to real data, the few existing ones are very promising. A potential path to boost progress in this subfield is to apply symbolic regression to experimental data in physics.

%% file: appendix.tex
\appendix

\section{Datasets Benchmarks Equations}
\label{app:benchmarks}

\begin{table}[htp]
    \caption{Ground-truth expressions for Koza~\cite{6791463}, Nguyen~\cite{Uy2010SemanticallybasedCI}, Jin~\cite{jin}, Keijzer~\cite{keijer} and R~\cite{R} benchmarks.}
    \label{tab:srprob1}
    \begin{tabular}{@{}llll@{}}
    \toprule
    Dataset & Expression & Variables & Data range\\
    \midrule
    Koza-1\footnote{Same as Nguyen-2} & $x^4 + x^3 + x^2 + x$ & 1 &  U[-1, 1, 20] \\
    Koza-2 & $x^5 - 2x^3 + x$ & 1 & U[-1, 1, 20] \\
    Koza-3 & $x^6 - 2x^4 + x^2$ & 1 & U[-1, 1, 20] \\
    \midrule
    Nguyen-1 & $x^3+ x^2 + x$ & 1 & U(-1,1,20)\\
    Nguyen-2 & $x^4 + x^3+ x^2 + x$& 1 &  U(-1,1,20)\\
    Nguyen-3 & $x^5 + x^4 + x^3+ x^2 + x$& 1 & U(-1,1,20)\\
    Nguyen-4 & $x^6 + x^5 + x^4 + x^3+ x^2 + x$ & 1 & U(-1,1,20)\\
    Nguyen-5 & $\sin(x^2)\cos(x) -1$ & 1 & U(-1,1,20)\\
    Nguyen-6 & $\sin(x) + \sin(x+x^2)$ & 1 & U(-1,1,20)\\
    Nguyen-7 & $\log(x+1) + \log(x^2+1)$ & 1 & U(0,2,20)\\
    Nguyen-8 & $\sqrt{x}$ & 1 & U(0,4,20)\\
    Nguyen-9 & $\sin(x) + \sin(y^2)$ & 2 & U(-1,1,100)\\
    Nguyen-10 & $2\sin(x)\cos(y)$ & 2 & U(-1,1,100)\\
    Nguyen-11 & $x^{y}$& 2\\
    Nguyen-12 & $x^4 - x^3 + \frac{1}{2}y^2 - y$& 2\\
    \midrule
    Jin-1 & $2.5x^4 -1.3x^3 +0.5y^2 -1.7y$  & 2 & U(-3,3,100)\\
    Jin-2 & $ 8.0x^2 + 8.0y^3 -15.0$ & 2 & U(-3,3,100)\\
    Jin-3 & $ 0.2x^3 +1.5y^3 -1.2y -0.5x$ & 2 & U(-3,3,100)\\
    Jin-4 & $ 1.5\exp(x) + 5.0\cos(y)$ & 2 & U(-3,3,100)\\
    Jin-5 & $ 6.0\sin(x)\cos(y)$ & 2 & U(-3,3,100)\\
    Jin-6 & $ 1.35xy + 5.5\sin((x-1.0)(y-1.0)$ & 2 & U(-3,3,100)\\
    \midrule
    Keijzer-1 & $0.3 x \sin(2\pi x)$ & 1 & E[-1, 1, 0.1] \\
    Keijzer-2 & $0.3 x \sin(2\pi x)$ & 1 & E[-2, 2, 0.1] \\
    Keijzer-3 & $0.3 x \sin(2\pi x)$ & 1 & E[-3, 3, 0.1] \\
    Keijzer-4 & $x^3e^{-x} \cos(x)\sin(x)(\sin^2(x)\cos(x)-1)$ & 1 & E[0, 10, 0.05] \\
    \multirow{2}{*}{Keijzer-5} & \multirow{2}{*}{$30xz/(x-10)y^2$} & \multirow{2}{*}{3} & $x,z:$ U[-1,1,1000] \\ 
    & & & $y:$ U[1,2,1000]\\
    Keijzer-6 & $\sum_1^{x}i$ & 1 & E[1, 50, 1] \\
    Keijzer-7 & $\log x$ & 1 & E[1, 100, 1] \\
    Keijzer-8 & $\sqrt{x}$ & 1 & E[0, 100, 1] \\
    Keijzer-9 & $\mathrm{arcsinh}(x)=\log(x +\sqrt{x^2 + 1})$ & 1 & E[0, 100, 1] \\
    Keijzer-10 & $x^y$ & 2 & U[0, 1, 100] \\
    Keijzer-11 & $xy + \sin((x-1)(y-1))$ & 2 & U[-3, 3, 20] \\
    Keijzer-12 & $x^4-x^3 +y^2/2 - y$ & 2 & U[-3, 3, 20] \\
    Keijzer-13 & $6\sin(x)\cos(y)$ & 2 & U[-3, 3, 20] \\
    Keijzer-14 & $8/(2+x^2+y^2)$ & 2 &  U[-3, 3, 20] \\
    Keijzer-15 & $x^3/5 +y^3/2-y-x$ & 2 & U[-3, 3, 20] \\
    \midrule
    R1 & $(x+1)^3/(x^2-x+1)$ & 1 & E[-1,1,20]\\
    R2 & $(x^5-3x^3+1)/(x^2+1)$ & 1 & E[-1,1,20]\\
    R3 & $(x^6+x^5)/(x^4+x^3+x^2+x+1)$ & 1 & E[-1,1,20]\\
    \botrule
\end{tabular}
\end{table}

\begin{table}[htp]
    \caption{Ground-truth expressions for  Korns~\cite{Korns2011} and Livermore~\cite{dsr} benchmarks.}
    \label{tab:srprob2}
    \begin{tabular}{@{}llll@{}}
    \toprule
    Dataset & Expression & Variables & Data range\\
    \midrule
    Korns-1 & $1.57 + (24.3 v)$ & 1 & U[-50, 50, 10000] \\
    Korns-2 & $0.23 + 14.2\frac{v+y}{3\omega}$ & 3 & U[-50, 50, 10000] \\
    Korns-3 & $-5.41 + 4.9\frac{v-x+y/w}{3\omega}$ & 4 & U[-50, 50, 10000]\\
    Korns-4 & $-2.3 + 0.13\sin(z)$ & 1 & U[-50, 50, 10000] \\
    Korns-5 & $3 + 2.13 \ln(\omega)$ & 1 & U[-50, 50, 10000] \\
    Korns-6 & $1.3 + 0.13 \sqrt{x}$ & 1 &  U[-50, 50, 10000] \\
    Korns-7 & $213.80940889(1- e^{-0.54723748542 x})$ & 1 & U[-50, 50, 10000] \\
    Korns-8 & $6.87 + 11 \sqrt{7.23~x~v~\omega}$ & 3 &  U[-50, 50, 10000] \\
    Korns-9 & $\frac{\sqrt{x}}{\ln(y)}\frac{e^z}{v^2}$ & 4 &  U[-50, 50, 10000] \\[5pt]
    Korns-10 & $0.81 + 24.3\frac{2 y+3 z^2}{4v^3+5\omega^4}$ & 4 & U[-50, 50, 10000] \\[5pt]
    Korns-11 & $6.87 + 11\cos(7.23 x^3)$ & 1 &  U[-50, 50, 10000] \\
    Korns-12 & $2-2.1\cos(9.8 x)\sin(1.3\omega)$ & 2 & U[-50, 50, 10000] \\
    Korns-13 & $32-3\frac{\tan(x)}{\tan(y)}\frac{\tan(z)}{\tan(v)}$ & 4 & U[-50, 50, 10000]\\
    Korns-14 & $22-4.2(\cos(x)-\tan(y))\frac{\tanh(z)}{\sin(v)}$ & 4 & U[-50, 50, 10000] \\
    Korns-15 & $12-6\frac{\tan(x)}{e^y}(\ln(z)-\tan(v))$ & 4 &  U[-50, 50, 10000]\\
    \midrule
    Livermore-1 & $1/3 + x + \sin(x^2)$ & 1 & U[-10,10,1000] \\
    Livermore-2 & $\sin(x^2)\cos(x) - 2$ & 1 & U[-1,1,20]\\
    Livermore-3 & $\sin(x^3)\cos(x^2) -1$ & 1 & U[-1,1,20]\\
    Livermore-4 & $\log(x+1) + \log(x^2+1)+\log(x)$ & 1 & U[0,2,20]\\
    Livermore-5 & $x^4 - x^3 + x^2 -y$ & 2 & U[0,1,20]\\
    Livermore-6 & $4x^4 + 3x^3 + 2x^2 + x$ & 1 & U[-1,1,20]\\
    Livermore-7 & $\sinh(x)$ & 1 & U[-1,1,20]\\
    Livermore-8 & $\cosh(x)$ & 1 & U[-1,1,20]\\
    Livermore-9 & $x^9 +x^8+x^7+x^6+x^5+x^4+x^3+x^2+x$ & 1 & U[-1,1,20]\\
    Livermore-10 & $6\sin(x)\cos(y)$ & 2 & U[0,1,20]\\
    Livermore-11 & $x^2y^2/(x+y)$ & 2 & U[-1,1,50]\\
    Livermore-12 & $x^5/y^3$ & 2 & U[-1,1,50]\\
    Livermore-13 & $x^{1/3}$ & 1 & U[0,4,20]\\
    Livermore-14 & $x^3+x^2+x+\sin(x)+\sin(x^2)$ & 1 & U[-1,1,20]\\
    Livermore-15 & $x^{1/5}$ & 1 & U[0,4,20]\\
    Livermore-16 & $x^{2/5}$ & 1 & U[0,4,20]\\
    Livermore-17 & $4\sin(x)\cos(y)$ & 2 & U[0,1,20]\\
    Livermore-18 & $\sin(x^2)\cos(x) - 5$ & 1 & U[-1,1,20]\\
    Livermore-19 & $x^5+x^4+x^2+x$ & 1 & U[-1,1,20]\\
    Livermore-20 & $\exp(-x^2)$ & 1 & U[-1,1,20]\\
    Livermore-21 & $x^8+x^7+x^6+x^5+x^4+x^3+x^2+x$ & 1 & U[-1,1,20]\\
    Livermore-22 & $\exp(-0.5x^2)$ & 1 & U[-1,1,20]\\
    \botrule
\end{tabular}
\end{table}

\begin{table}[htp]
    \caption{Ground-truth expressions for Vladislavleva~\cite{vladislavleva} benchmark.}
    \label{tab:srprob3}
    \begin{tabular}{@{}llll@{}}
    \toprule
    Dataset & Expression & Variables & Data range\\
    \midrule
    Vladislavleva-1 & $\frac{e^{-(x-1)^2}}{1.2+(y-2.5)^2}$ & 1 & U[0.3, 4, 100] \\
    Vladislavleva-2 & $e^{-x}x^3(\cos x\sin x)(\cos x \sin^2 x-1)$ & 2 & E[0.5, 10, 0.1] \\
    Vladislavleva-3 & $e^{-x}x^3(\cos x\sin x)(\cos x\sin^2 x-1)(y-5)$ & 2 & $x:$E[0.05,10,0.1]\\
    & & & $y:$E[0.05,10.05,2]\\
    Vladislavleva-4 & $\frac{10}{5+\sum_{i=1}^{5}(x_i-3)^2}$ & 5 & U[0.05, 6.05, 1024] \\
    Vladislavleva-5 & $30(x-1)\frac{(z-1)}{y^2(x-10)}$ & 3 & $x:$ U[0.05, 2, 300]\\ 
    & & & $y:$ U[1, 2, 300] \\
    & & & $z:$ U[0.05, 2, 300] \\
    Vladislavleva-6 & $6\sin(x)\cos(y)$ & 2 & U[0.1, 5.9, 30] \\
    Vladislavleva-7 & $(x-3)(y-3) + 2\sin((x-4)(y-4))$ & 2 & U[0.05, 6.05, 300] \\
    Vladislavleva-8 & $\frac{(x-3)^4+(y-3)^3-(y-3)}{(y-2)^4+10}$ & 2 &  U[0.05, 6.05, 50]\\ 
    \botrule
\end{tabular}
\end{table}

\begin{table}[!htb]
    \caption{Feynman physics equation~\cite{Udrescu:2019mnk}.}
    \label{tab:aifeynman1}
    \begin{minipage}{.5\linewidth}
      \centering
        \begin{tabular}{@{}ll@{}}
        \toprule
        Function form & \#v\\
        \toprule
        $f=\exp(-\theta^2/2)/\sqrt(2\pi)$&1\\
        $f=\exp(-(\theta/\sigma)^2/2)/(\sqrt(2\pi)\sigma)$&2\\
        $f=\exp(-((\theta-\theta_1)/\sigma)^2/2)/(\sqrt(2\pi)\sigma)$&3\\
        $d=\sqrt{(x_2-x_1)^2+(y_2-y_1)^2}$&4\\
        $F=\frac{Gm1m2}{((x_2-x_1)^2+(y_2-y_1)^2+(z_2-z_1)^2}$&9\\
        $m=\frac{m_0}{\sqrt{1-v^2/c^2}}$&3\\
        $A=x_1y_1 + x_2y_2 + x_3y_3$&6\\
        $F=\mu N_n$&2\\
        $F=q_1q_2/(4\pi\epsilon r^2)$&4\\
        $E_f=q_1 r/(4\pi\epsilon r^3)$&3\\
        $F=q_2E_f$&2\\
        $F=q(E_f+Bv\sin(\theta))$&5\\
        $K=1/2m(v^2+u^2+w^2)$&4\\
        $U=Gm1m2(\frac{1}{r_2}-\frac{1}{r_1})$&5\\
        $U=mgz$&3\\
        $U=\frac{1}{2}k_{spring} x^2$&2\\
        $x^{\prime}=(x-ut)/\sqrt{1-u^2/c^2}$&4\\
        $t^{\prime}=(t-ux/c^2)/\sqrt{1-u^2/c^2}$&4\\
        $p=m_0v/\sqrt{1-v^2/c^2}$&3\\
        $v^{\prime}=(u+v)/(1+uv/c^2)$&3\\
        $r=(m_1r_1+m_2r_2)/(m1+m2)$&4\\
        $\tau=rF\sin(\theta)$&3\\
        $L=mrv\sin(\theta)$&4\\
        $E=\frac{1}{4}m(\omega^2+\omega_0^2)x^2$&4\\
        $V_e=q/C$&2\\
        $\theta_1=\arcsin(n\sin(\theta2))$&2\\
        $f_f=1/(\frac{1}{d_1}+\frac{n}{d_2})$&3\\
        $k=\omega/c$&2\\
        $x=\sqrt{x_1^2+x_2^2-2x_1x_2\cos(\theta_1-\theta_2)}$&4\\
        $I_{\star}=I_{0\star}\sin^2(n\theta/2)/\sin^2(\theta/2)$&3\\
        $\theta=\arcsin(\lambda/nd)$&3\\
        $P=q^2a^2/(6\pi\epsilon c^3)$&4\\
        $P=(1/2\epsilon cE_f^2)(8\pi r^2/3)(\omega^4/(\omega^2-\omega_0^2)^2)$&6\\
        $\omega=qvB/p$&4\\
        $\omega=\omega_0/(1-v/c)$&3\\
        $\omega=(1+v/c)/\sqrt{1-v^2/c^2}\omega_0$&3\\
        $E=\bar{h}\omega$&2\\
        $I_{\star}=I_1+I_2+2\sqrt{I_1I_2}\cos(\delta)$&3\\
        $r = 4\pi\epsilon\bar{h}^2/(mq^2)$&4\\
        $E=\frac{3}{2}p_FV$&2\\
        $E=1/(\gamma-1)p_FV$&3\\
        $P_F=nkb_T/V$&4\\
        \midrule
        \end{tabular}
    \end{minipage}%
    \begin{minipage}{.5\linewidth}
      \centering
        \begin{tabular}{@{}ll@{}}
        \toprule
        Function form & \#v\\
        \toprule
        $n=n_0\exp(-mgx/(k_bT))$&6\\
        $L_{rad}=\bar{h}\omega^3/(\pi^2c^2(\exp(\bar{h}\omega/(k_bT))-1))$&5\\
        $v=mu_{drift}qV_e/d$&4\\
        $D=\mu_e k_bT$&3\\
        $\kappa=1/(\gamma-1)k_bv/A$&4\\
        $E=nk_bT\ln(\frac{V2}{V1})$&5\\
        $c=\sqrt(\gamma pr/\rho)$&3\\
        $E = mc^2/\sqrt{1-v^2/c^2}$&3\\
        $x = x_1(\cos(\omega t)+\alpha\cos(\omega t)^2)$&4\\
        $P = \kappa(T_2-T_1)A/d$&5\\
        $F_E=Pwr/(4\pi r^2)$&2\\
        $V_e=q/(4\pi\epsilon r)$&3\\
        $V_e=\frac{1}{4\pi\epsilon}p_d\cos(\theta)/r^2$&4\\
        $E_f=\frac{3}{4\pi\epsilon}p_dz/r^5\sqrt{x^2+y^2}$&6\\
        $E_f=\frac{3}{4\pi\epsilon}p_d\cos(\theta)\sin(\theta)/r^3$&4\\
        $E=\frac{3}{5}q^2/(4\pi\epsilon d)$&3\\
        $E_{den}=\epsilon Ef^2/2$&2\\
        $E_f=\sigma_{den}/\epsilon 1/(1+\chi)$&3\\
        $x=qE_f/(m(\omega_0^2-\omega^2))$&5\\
        $n=n_0(1+p_dE_f\cos(\theta)/(k_bT))$&6\\
        $P_{\star}=n_{rho}p_d^2E_f/(3k_bT)$&5\\
        $P_{\star}=n\alpha/(1-(n\alpha/3))\epsilon E_f$&4\\
        $\theta=1+n\alpha/(1-(n\alpha/3))$&2\\
        $B=1/(4\pi\epsilon c^2)2I/r$&4\\
        $\rho_{c}=\rho_{c_0}/\sqrt{1-v^2/c^2}$&3\\
        $j=\rho_{c_0}v/\sqrt{1-v^2/c^2}$&3\\
        $E=-\mu_{M}B\cos(\theta)$&3\\
        $E=-p_dE_f\cos(\theta)$&3\\
        $V_e=q/(4\pi\epsilon r(1-v/c))$&5\\
        $k=\sqrt{\omega^2/c^2-\pi^2/d^2}$&3\\
        $F_E=\epsilon cE_f^2$&3\\
        $E_{den}=\epsilon E_f^2$&2\\
        $I=qv/(2\pi r)$&3\\
        $\mu_{M}=qvr/2$&3\\
        $\omega=gqB/(2m)$&4\\
        $\mu_{M}=q\bar{h}/(2 m)$&3\\
        $E=g\mu_{M}BJ_z/\bar{h}$&5\\
        $M=n_{rho}\mu_{M}\tanh(\mu_{M}B/(k_bT))$&5\\
        $f=\mu_m B/(k_bT)+(\mu_m\alpha)/(\epsilon c^2k_bT)M$&8\\
        $E=\mu_{M}(1+\chi)B$&6\\
        $F=YA_x/d$&4\\
        $\mu_S=Y/(2(1+\sigma))$&2\\
        $E=\bar{h}\omega/(\exp(\bar{h}\omega/(k_bT))-1)$&4\\
        \midrule
        \end{tabular}
    \end{minipage} 
\end{table}

\begin{table}[htp]
\begin{center}
\caption{Feynman physics equation~\cite{Udrescu:2019mnk}.}
\label{tab:aifeynman3}
\begin{tabular}{@{}ll@{}}
\toprule
Function form & \# variables \\
\toprule
$n=1/(\exp(\bar{h}\omega/(k_bT))-1)$&4\\
$n=n_0/(\exp(\mu_m B/(k_bT))+\exp(-\mu_m B/(k_bT)))$\\
$\omega=2\mu_M B/\bar{h}$&3\\
$p_{\gamma}=\sin(E_n t/\bar{h})^2$&3\\
$p_{\gamma}=(p_dE_ft/\bar{h})\sin((\omega-\omega_0)t/2)^2/((\omega-\omega_0)t/2)^2$&6\\
$E=\mu_M\sqrt{B_x^2+B_y^2+B_z^2}$&3\\
$L=n\bar{h}$&2\\
$v=2E_nd^2k/\bar{h}$&4\\
$I=I_0(\exp(qV_e/(k_bT))-1)$&5\\
$E=2U(1-\cos(kd))$&3\\
$m=\bar{h}^2/(2E_nd^2)$&3\\
$k=2\pi\alpha/(nd)$&3\\
$f=\beta(1+\alpha\cos(\theta))$&3\\
$E=-mq^4/(2(4\pi\epsilon)^2\bar{h}^2)(1/n^2)$&4\\
$j=-\rho_{c_0}qA_{vec}/m$&4\\
\midrule
\end{tabular}
\end{center}
\end{table}